\documentclass{article}

\PassOptionsToPackage{numbers, compress}{natbib}

 \usepackage[preprint]{neurips_2026}

\usepackage[utf8]{inputenc} 
\usepackage[T1]{fontenc}    
\usepackage{url}            
\usepackage{booktabs}       
\usepackage{amsfonts}       
\usepackage{nicefrac}       
\usepackage{microtype}      

\usepackage{colortbl}
\usepackage{makecell}
\usepackage{arydshln}

\usepackage{soul}
\usepackage[colorlinks,
            linkcolor=blue,       
            anchorcolor=blue,  
            citecolor=blue,        
            ]{hyperref}

\usepackage{amsthm}
\usepackage{amsmath}        
\usepackage{graphicx}       
\usepackage{float}          
\usepackage{subcaption}     
\usepackage[noend]{algpseudocode}   
\usepackage[ruled]{algorithm2e}
\usepackage{amsfonts,amssymb}  
\usepackage{mathrsfs}       
\usepackage{bm}             
\usepackage{circledsteps}   
\usepackage{stfloats}
\usepackage{array}          
\usepackage{arydshln}
\usepackage{tabularx}       

\usepackage[table]{xcolor}
\usepackage{booktabs}
\usepackage{multirow}

\usepackage{wrapfig}

\usepackage{subcaption}
\usepackage{xspace}

\definecolor{ElegantPurple}{RGB}{102,8,116}

\newcommand{\sysname}[0]{\textcolor{black}{ElegantVLA}\xspace}

\title{ElegantVLA: Learning When to Think for Efficient Vision-Language-Action Models}

\author{
    \textbf{Ye Li}$^{1}$,
    \textbf{Huanan Liu}$^{1}$,
    \textbf{Kangye Ji}$^{1}$,
    \textbf{Yuan Meng}$^{1}$\thanks{Corresponding Authors: yuanmeng@tsinghua.edu.cn, wangzhi@sz.tsinghua.edu.cn.},
    \textbf{Jiajun Fan}$^{2}$,
    \textbf{Yuansong Wang}$^{1}$,
    \textbf{Shiyu Qin}$^{1}$, \\
    \textbf{Chenglei Wu}$^{1}$, 
    \textbf{Shu-Tao Xia}$^{1}$,
    \textbf{Zhi Wang}$^{1\ast}$, \\
    $^1$ Tsinghua University \quad
    $^2$ University of Illinois at Urbana-Champaign \quad
}

\begin{document}

\maketitle


\begin{abstract}
Vision-Language-Action (VLA) models have emerged as a powerful paradigm for generalist robotic control. 
However, their high computational cost and limited control frequency hinder real-time robotic manipulation, especially when large vision-language backbones and iterative action heads are executed at every control step. 
Existing VLA acceleration methods often optimize individual components or rely on fixed acceleration rules, treating different control steps with largely fixed computation and overlooking the non-uniform reasoning demands of sequential embodied control. 
Inspired by human motor control, where cognitive and feedback resources concentrate on goal-sensitive stages, we argue that VLA models should learn when to invest full computation and when to reuse prior computation. 
To this end, we propose \textbf{ElegantVLA}, a plug-in phase-adaptive inference framework that accelerates VLA models through intra-model dynamic compute scheduling. 
ElegantVLA introduces a lightweight scheduler that observes temporal representation similarity, robot-motion cues, and episode progress to jointly allocate computation across the vision encoder, LLM, and action head. 
For perception-language reasoning, the scheduler selects a five-level Vision--LLM compute mode, from full recomputation to multi-step temporal reuse, based on visual-language representation stability.
For action generation, it selects a three-level denoising mode, reusing intermediate denoising states during stable motion while preserving full refinement for goal-sensitive stages. 
By coordinating these decisions, ElegantVLA provides a general acceleration framework for modern VLA pipelines with explicit action-generation modules, without modifying or retraining the base model. 
Extensive experiments on GR00T, CogACT, and real-world tasks show that ElegantVLA preserves or improves task success while substantially accelerating inference.
On GR00T, it achieves up to 2.55$\times$ average speedup.
On CogACT, it delivers a 3.77$\times$ average speedup.
In GR00T-based real-world experiments across six tasks, it reduces computation by 2.18$\times$ and increases control frequency from 13.8 Hz to 26.3 Hz.
The project website is available at \href{https://anonymous.4open.science/w/elegantvla/}{Project Page}.

\end{abstract}

\section{Introduction}

Vision-Language-Action (VLA) models have emerged as a powerful paradigm for generalist robotic control by connecting visual perception, language understanding, and action generation in a unified policy \citep{zitkovich2023rt,kim2024openvla,team2024octo,black2024pi_0,li2024cogact,liu2024rdt}.
Recent systems have shown impressive generalization across manipulation tasks and environments, but this capability comes with a substantial inference cost.
Modern VLA pipelines typically combine a vision encoder, a large language model (LLM), and an action-generation module such as an autoregressive decoder, flow model, or diffusion head.
Executing all these components at every control step limits the control frequency of VLA policies, making real-time deployment difficult in latency-sensitive robotic manipulation.

Existing acceleration methods have made progress across model design, perception, action generation, and inference optimization \citep{guan2025efficient,yu2025survey}.
Some works reduce VLA computation through early exit, quantization, layer skipping, or adaptive computation routing \citep{yue2024deer,park2024quantization,yang2025efficientvla,yang2026dysl,zheng2026dyq,yu2026ac,li2025prance}.
Others exploit visual redundancy through token pruning, token caching, token fusion, or action-aware visual selection \citep{xu2025vla,li2025sp,tan2025think,wang2025specprune,pei2025action,liu2025vla,jiang2025better}.
Recent work also accelerates action generation through diffusion-policy caching, parallel decoding, speculative decoding, and flow-specific inference design \citep{li2025ts,ji2025block,song2025accelerating,lu2026faster}.
However, most existing methods still optimize a specific component, or rely on a fixed acceleration rule, while the whole VLA pipeline is repeatedly invoked under largely uniform computation across control steps.
This overlooks a key property of robotic control: 
the computational demands of perception, reasoning, and action generation vary significantly across different task phases.

This non-uniformity is also natural from the perspective of human motor control.
Human movements are not controlled with the same cognitive and feedback effort throughout an entire task.
Stable transport phases can rely more on prediction and low-level feedback, while goal-sensitive phases such as contact, alignment, grasping, insertion, and placement require more careful correction and control \citep{wolpert1995internal,harris1998signal,todorov2002optimal}.
VLA policies exhibit an analogous structure during sequential interaction.
Adjacent control steps often share similar visual observations, semantic context, and action intent, whereas goal-sensitive stages may induce abrupt representation changes and require full effort.

\begin{figure}[t]
    \centering
    \includegraphics[width=\linewidth]{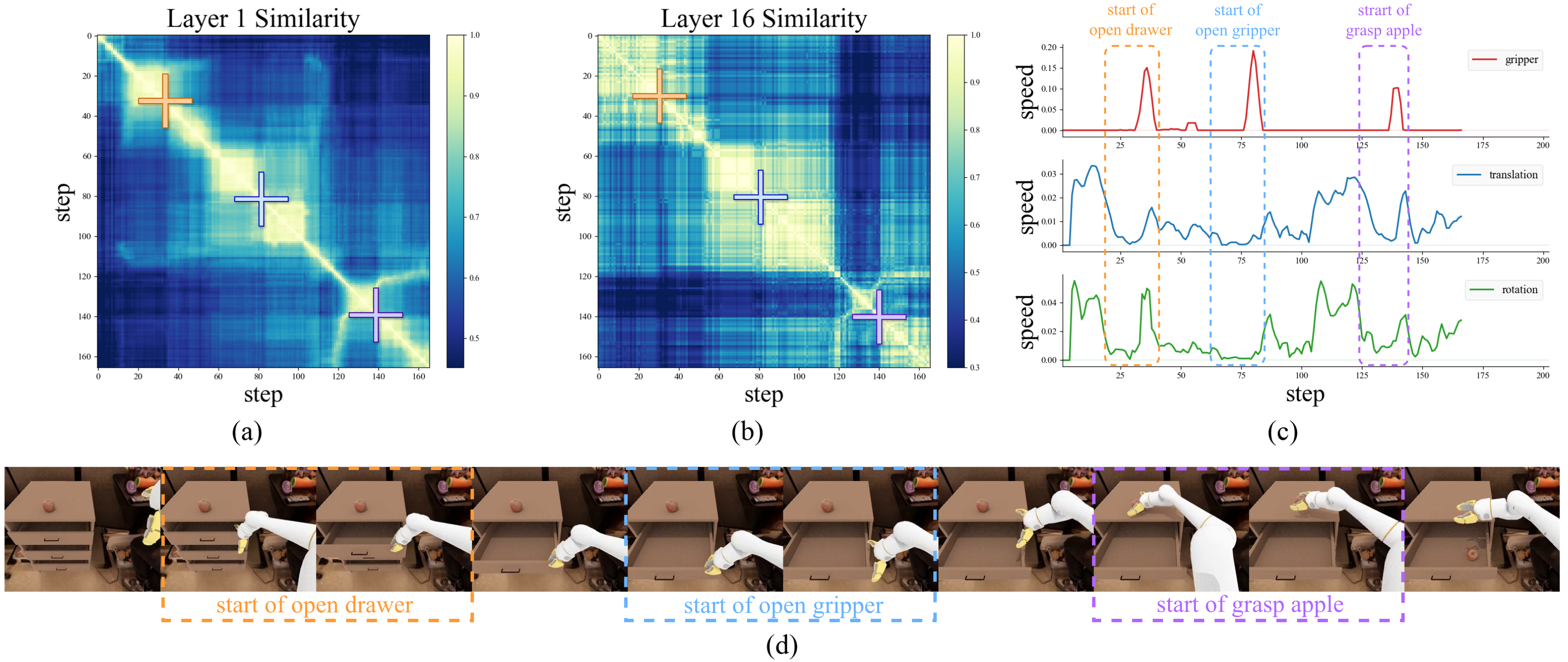}
    \caption{
        \textbf{Observation motivating phase-adaptive cache scheduling.}
    VLA execution exhibits temporally varying inference demand rather than uniform step-wise complexity.
    In this drawer-opening and apple-grasping rollout, (a,b) show pairwise CKA similarities of the first and final LLM-layer representations, (c) shows robot motion speeds, and (d) shows the rollout.
    Similar changes appear in both final-layer and first-layer similarities, suggesting that a lightweight first-layer probe can inform cache/recompute scheduling.
    }
    \label{fig:observation}
\end{figure}

Therefore, in this work, we propose a phase-adaptive VLA acceleration method to further reduce computational redundancy during inference.
This design raises two core challenges.
First, how can a VLA system estimate the per-step per-module computational demand without explicit phase labels?
The computational demand of each control step is not explicitly given and changes continuously as the task evolves, making it difficult to decide when full inference is necessary.
Second, how can computation be coordinated across the full VLA pipeline without degrading task success?
Many existing methods make acceleration decisions at the level of individual modules, or combine several component-wise rules, without explicitly modeling how these decisions interact across the full perception-reasoning-action pipeline over time.
As a result, locally aggressive reuse may save computation immediately, but can introduce temporal errors that propagate and accumulate during task execution.
An effective method must therefore coordinate reuse decisions across modules while accounting for their long-term effects on control stability.

To address these challenges, we propose \sysname, a plug-in phase-adaptive inference framework for efficient VLA models.
First, we design multi-level computation strategies for the perception-reasoning and action generation modules respectively. 
A lightweight scheduler then determines the computation strategy for each control step based on task-relevant temporal representation similarity and robot-motion cues.
For perception-language reasoning, the scheduler selects a five-level Vision--LLM compute mode, from full recomputation to multi-step temporal reuse, based on visual-language representation stability.
For action generation, \sysname selects a three-level denoising mode, reusing intermediate denoising states during stable motion while preserving full refinement for precision-sensitive stages.
In this way, \sysname turns VLA acceleration into an intra-model dynamic compute scheduling problem, rather than a static compression or single-module pruning problem.
Second, we introduce reinforcement learning (RL) to train the scheduler for globally optimal computation allocation. We formulate scheduling as a sequential decision problem, where each decision trades off task performance, computation cost, and reuse stability.
The RL policy is conditioned on interpretable scheduling signals, including CKA-based temporal representation similarity, robot motion speeds, and episode progress, while its training objective encourages efficient reuse without allowing errors to accumulate across control steps.
This enables \sysname to coordinate recomputation and reuse across perception, reasoning, and action generation under one policy.

We validate \sysname in simulation and real-world manipulation.
On GR00T in SimplerEnv, \sysname improves success rate from 64.00\% to 65.88\% while achieving up to 2.55$\times$ average FLOPs speedup.
On CogACT, it reaches 3.72$\times$ and 3.77$\times$ average FLOPs speedups on Visual Matching and Variant Aggregation, respectively.
In GR00T-based real-world experiments across six tasks, \sysname reduces computation by 2.18$\times$ and increases control frequency from 13.8 Hz to 26.3 Hz, while improving average success from 61.67\% to 65.00\%.
The improvement is visible on conveyor-belt pickup tasks, where faster control helps the policy react to moving targets and improves success on pineapple bun and toast.
The main contributions are as follows:
\begin{itemize}
\item [(1)] We formulate efficient VLA inference as a phase-adaptive intra-model compute allocation problem.
Inspired by human motor control, we argue that computational effort should vary with task progress, semantic change, and robot state rather than remaining fixed across control steps.
\item [(2)] We propose \sysname, a unified temporal reuse framework for VLA acceleration.
\sysname dynamically coordinates computation across the vision encoder, LLM, and action head through full computation, partial recomputation, and temporal reuse.
\item [(3)] We develop a scheduler learning framework tailored to phase-adaptive VLA inference.
The scheduler learns stable recomputation and reuse decisions by balancing task performance, compute efficiency, and reuse stability.
\end{itemize}

\section{Related Work}

\textbf{Vision Language Action Models.}
Representative systems such as OpenVLA, GROOT, $\pi_0$, and RDT have demonstrated strong generalization across manipulation settings by combining semantic perception, language-conditioned reasoning, and action generation \citep{zitkovich2023rt,kim2024openvla,bjorck2025gr00t,black2024pi_0,li2024cogact,liu2024rdt}.
Recent VLA architectures increasingly separate perception-language reasoning from action generation: a vision encoder and LLM produce task-conditioned representations, while an action decoder, diffusion policy, or flow-based action head generates continuous control.
This modular design improves the expressiveness and flexibility of VLA policies, but it also distributes inference cost across the full perception-reasoning-action pipeline.
Thus, efficient VLA deployment requires understanding computation across perception, reasoning, and action generation, rather than accelerating a single backbone.

\textbf{Efficient VLA Inference.}
Recent work has explored VLA acceleration through model-level compression and routing, perception-side token compression, and faster action generation \citep{guan2025efficient,yu2025survey,yue2024deer,park2024quantization,yang2025efficientvla,yang2026dysl,zheng2026dyq,yu2026ac,xu2025vla,li2025sp,tan2025think,wang2025specprune,li2023ddpg,liu2024novel,pei2025action,liu2025vla,jiang2025better,liu2026ttf}.
Another line targets diffusion-based action heads with block-wise caching, sparse denoising, parallel decoding, or flow-specific inference design \citep{li2025ts,ji2025block,ji2026sparse,song2025accelerating,lu2026faster}.
These methods reveal substantial redundancy in VLA inference, but they often optimize a single module or rely on component-wise rules.
More broadly, existing methods rarely model how computation demand evolves during execution or how cross-module reuse interacts over long horizons.
Inspired by human motor control, where prediction, feedback correction, and precision-dependent control are allocated non-uniformly over time \citep{wolpert1995internal,harris1998signal,todorov2002optimal}, \sysname learns a phase-adaptive scheduler that coordinates recomputation and reuse across the vision encoder, LLM, and action head.

\section{Method}
\label{sec:method}

\begin{figure}[t]
    \centering
    \includegraphics[width=\linewidth]{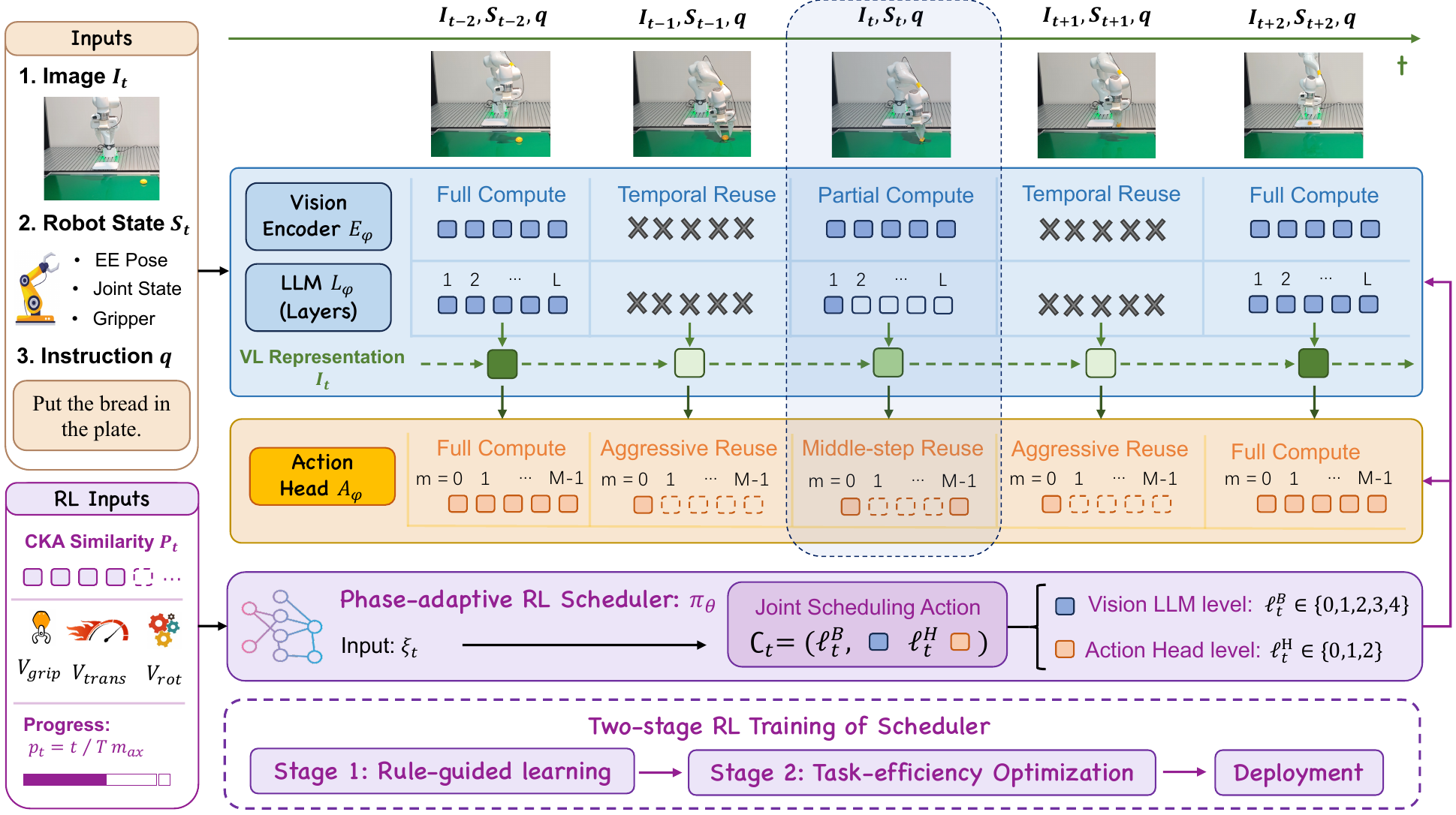}
    \caption{
        \textbf{Overview of \sysname.}
        \sysname learns when to think during VLA control by attaching a phase-adaptive scheduler to a frozen base policy.
        At each step, the scheduler uses temporal representation similarity, robot-motion signals, and episode progress to select a joint compute action for the Vision-LLM backbone and the action head.
        It allocates full recomputation to phase-sensitive moments and reuses cached computation during stable phases, enabling efficient inference without modifying or retraining the base VLA model.
    }
    \label{fig:framework}
\end{figure}

\subsection{Phase-Adaptive Scheduling Framework}
\label{sec:method_overview}

\sysname formulates efficient VLA inference as a temporal compute allocation problem for embodied control.
This view is motivated by human motor control, where prediction and feedback are allocated unevenly across movement phases rather than applied uniformly at every instant \citep{wolpert1995internal,harris1998signal,todorov2002optimal}.
Analogously, a VLA policy need not spend the same amount of perception-language reasoning and action refinement at every control step.
We instantiate this view as a scheduler-based inference wrapper for frozen VLA policies, operating on two branches: the \emph{Vision-LLM backbone}, which contains the vision encoder and LLM, and the \emph{action head}, which converts visual-language representations and robot states into executable actions.
Given an image observation $\mathbf{I}_t$, robot state $\mathbf{s}_t$, and language instruction $q$, a standard VLA policy performs
\begin{equation}
    \mathbf{z}_t = \mathcal{L}_\phi(\mathcal{E}_\phi(\mathbf{I}_t), q), \qquad
    \hat{\mathbf{a}}_{t:t+K-1} = \mathcal{H}_\phi(\mathbf{z}_t, \mathbf{s}_t),
\end{equation}
where $\mathcal{E}_\phi$, $\mathcal{L}_\phi$, and $\mathcal{H}_\phi$ are kept fixed.
\sysname adds a lightweight scheduler $\pi_\theta$ that selects how to execute the Vision-LLM backbone and action head at each control step:
\begin{equation}
    \mathbf{c}_t=(\ell_t^{\mathrm{B}},\ell_t^{\mathrm{H}})\sim \pi_\theta(\cdot \mid \boldsymbol{\xi}_t).
\end{equation}
Here $\ell_t^{\mathrm{B}}$ controls the computation level of the Vision-LLM backbone, $\ell_t^{\mathrm{H}}$ controls how much action-head refinement is recomputed, and $\boldsymbol{\xi}_t$ is a compact observation built from temporal representation similarity and robot motion.
The Vision-LLM decision is organized as a five-level compute ladder, ranging from full recomputation to multi-step temporal reuse.
The action-head decision follows a three-level refinement-reuse ladder over iterative action generation: full refinement, middle-step reuse, and reuse after the first refinement step.
Optimizing the action head separately enlarges the scheduling space, allowing \sysname to preserve action precision when needed while still exploiting reuse for faster generation.
Together, these decisions make VLA inference better aligned with movement regularities, enabling task-complexity-aware temporal variable-frequency acceleration under a target accuracy constraint.
Figure~\ref{fig:framework} provides an overview of how \sysname couples backbone-level recomputation and action-head reuse through a learned phase-adaptive scheduler.

\subsection{Semantic-Stability-Guided Vision--LLM Scheduling}
\label{sec:backbone_scheduling}

The Vision-LLM backbone dominates per-step inference cost because it repeatedly encodes visual tokens and executes all LLM layers at the control frequency.
Following the motor-control view that prediction and feedback demands vary with movement phase \citep{wolpert1995internal,harris1998signal,todorov2002optimal}, stable motion require less frequent VLM reasoning than precision-sensitive phases.
\sysname therefore allocates backbone computation in a phase-adaptive manner by exposing five ordered compute levels to the scheduler.

As suggested by Figure~\ref{fig:observation}, temporal representation changes that appear in the final LLM layer are already visible after the first layer.
This motivates using first-layer CKA as a lightweight probe for backbone reuse decisions.
To estimate this stability, we use centered kernel alignment (CKA), which compares the similarity structure of hidden representations and is less tied to low-level pixel changes than raw input differences.
We compute CKA after the first LLM layer: at this point, visual tokens have interacted with the language instruction and the representation has been shaped by the LLM prior, making it a lightweight but semantically meaningful probe.
Specifically, \sysname maintains an anchor hidden-state matrix $\mathbf{H}_\tau$ from a recent full computation and compares it with the current first-layer hidden-state matrix $\mathbf{H}_t$:
\begin{equation}
    \rho_t = \operatorname{CKA}(\mathbf{H}_t, \mathbf{H}_\tau),
\end{equation}
where a larger $\rho_t$ indicates stronger temporal semantic stability.
The Vision-LLM action space contains five computation levels, $\ell_t^{\mathrm{B}} \in \{0,1,2,3,4\}$, ordered by increasing reuse aggressiveness.
When $\ell_t^{\mathrm{B}}=0$, \sysname computes both the vision encoder and the complete LLM, refreshing the anchor and output representation.
When $\ell_t^{\mathrm{B}}=1$, \sysname refreshes the vision encoder but only recomputes the first and last LLM layers, while reusing intermediate LLM representations from the most recent full computation.
This preserves input sensitivity and output alignment while avoiding the dominant cost of repeatedly executing all intermediate Transformer blocks.
When $\ell_t^{\mathrm{B}}=j$ for $j\in\{2,3,4\}$, \sysname skips both the vision encoder and the LLM for $j-1$ consecutive control steps and directly reuses the most recent visual-language representation.
\sysname feeds $\rho_t$ to the RL scheduler as one decision signal for selecting among these reuse levels.

Let $\bar{\mathbf{z}}$ denote the cached visual-language output representation from the most recent full or partial execution, and let $\bar{\mathbf{u}}$ denote the cached intermediate LLM states used by partial recomputation.
At a high level, the representation sent to the action head is
\begin{equation}
    \mathbf{z}_t =
    \begin{cases}
        \mathcal{L}_\phi(\mathcal{E}_\phi(\mathbf{I}_t), q), & \ell_t^{\mathrm{B}}=0,\\
        \mathcal{L}_\phi^{\mathrm{last}}(\bar{\mathbf{u}}, \mathcal{L}_\phi^{\mathrm{first}}(\mathcal{E}_\phi(\mathbf{I}_t), q)), & \ell_t^{\mathrm{B}}=1,\\
        \bar{\mathbf{z}}, & \ell_t^{\mathrm{B}}\in\{2,3,4\},
    \end{cases}
\end{equation}
This design aligns VLA inference with movement regularities and enables task-complexity-aware temporal variable-frequency acceleration while preserving optimal action accuracy.

\subsection{Motion-Aware Action-Head Scheduling}
\label{sec:action_head_scheduling}

Temporal redundancy in VLA inference appears at different levels.
The Vision-LLM backbone mainly captures high-level semantic redundancy in visual-language contexts, while the action head exhibits low-level redundancy in the iterative generation of continuous actions.
These two forms of redundancy are related but not identical: a stable semantic context may still require fresh action refinement due to changes in robot state or contact, whereas a refreshed visual-language representation may still lead to a smooth action-generation trajectory.
\sysname therefore schedules the action head as an independent branch, decoupling semantic reuse from action-generation precision.
Since action-refinement redundancy is tied to low-level motion continuity, \sysname exposes robot motion signals, including gripper speed, end-effector translation speed, and end-effector rotation speed, to the scheduler.
These signals help distinguish smooth transport phases, where refinement updates are more reusable, from precision-sensitive phases, where fresh action refinement is needed.
For an iterative action head with $M$ refinement steps,
\begin{equation}
    \mathbf{x}_t^{m+1}=\mathcal{F}_\phi^m(\mathbf{x}_t^m,\mathbf{z}_t,\mathbf{s}_t), \qquad m=0,\ldots,M-1,
\end{equation}
where $\mathcal{F}_\phi^m$ denotes the $m$-th refinement step inside the action head $\mathcal{H}_\phi$, and $\mathbf{x}_t^m$ is the intermediate action state.

\begin{table}[t]
  \centering
  \caption{
    \textbf{Simulation results on CogACT in SimplerEnv.}
    Each cell reports success rate (\%, $\uparrow$) / speedup ($\uparrow$).
    \sysname achieves 3.72$\times$ and 3.77$\times$ average speedup on Visual Matching and Variant Aggregation, respectively, while also obtaining the best average task success.
  }
  \label{tab:cogact_main_results}
  \begingroup
  \footnotesize
  \setlength{\tabcolsep}{2pt}
  \renewcommand{\arraystretch}{1.05}
  \begin{tabularx}{\linewidth}{l l *{5}{>{\centering\arraybackslash}X}}
    \toprule
    \multirow{2}{*}{\textbf{SIMPLER}} &
    \multirow{2}{*}{\textbf{Method}} &
    \multicolumn{4}{c}{\textbf{Success Rate (\%, $\uparrow$) / Speedup ($\uparrow$)}} &
    \multirow{2}{*}{\textbf{AVG}} \\
    \cmidrule(lr){3-6}
    & & PickCan & MoveNear & Drawer & PutDrawer & \\
    \midrule
    \multirow{6}{*}{\makecell{\textbf{Visual}\\\textbf{Matching}}}
    & CogACT~\citep{li2024cogact} & 91.30 / 1.00 & 85.00 / 1.00 & 71.80 / 1.00 & 50.90 / 1.00 & 74.80 / 1.00 \\
    \cmidrule(lr){2-7}
    & Random Dropping & 9.70 / 1.20 & 20.40 / 1.20 & 53.50 / 1.20 & 0.00 / 1.20 & 20.90 / 1.20 \\
    & FastV~\citep{chen2024image} & 82.67 / 1.31 & 72.92 / 1.31 & 66.67 / 1.28 & 46.30 / 1.30 & 67.14 / 1.30 \\
    & VLA-Cache~\citep{xu2025vla} & 87.33 / 1.47 & 77.92 / 1.46 & 69.44 / 1.44 & 47.22 / 1.47 & 70.48 / 1.46 \\
    & MoLe-VLA~\citep{zhang2025mole} & 86.40 / 1.53 & 80.20 / 1.53 & 70.60 / 1.53 & 50.40 / 1.53 & 71.90 / 1.53 \\
    & \cellcolor{ElegantPurple!15}\sysname & \cellcolor{ElegantPurple!15}\textbf{89.67} / \textbf{3.32} & \cellcolor{ElegantPurple!15}\textbf{80.42} / \textbf{3.05} & \cellcolor{ElegantPurple!15}\textbf{89.35} / \textbf{4.10} & \cellcolor{ElegantPurple!15}\textbf{50.93} / \textbf{4.41} & \cellcolor{ElegantPurple!15}\textbf{77.59} / \textbf{3.72} \\
    \midrule
    \multirow{6}{*}{\makecell{\textbf{Variant}\\\textbf{Aggregation}}}
    & CogACT~\citep{li2024cogact} & 89.60 / 1.00 & 80.80 / 1.00 & 28.30 / 1.00 & 46.60 / 1.00 & 61.30 / 1.00 \\
    \cmidrule(lr){2-7}
    & Random Dropping & 4.00 / 1.20 & 16.10 / 1.20 & 15.60 / 1.20 & 0.00 / 1.20 & 8.90 / 1.20 \\
    & FastV~\citep{chen2024image} & 76.00 / 1.29 & 71.67 / 1.29 & 24.34 / 1.26 & 47.62 / 1.26 & 54.91 / 1.28 \\
    & VLA-Cache~\citep{xu2025vla} & 80.00 / 1.47 & 75.00 / 1.45 & 30.69 / 1.43 & 43.39 / 1.43 & 57.27 / 1.45 \\
    & MoLe-VLA~\citep{zhang2025mole} & \textbf{89.20} / 1.49 & \textbf{79.50} / 1.49 & 29.90 / 1.49 & 46.20 / 1.49 & 61.20 / 1.49 \\
    & \cellcolor{ElegantPurple!15}\sysname & \cellcolor{ElegantPurple!15}82.30 / \textbf{3.32} & \cellcolor{ElegantPurple!15}77.17 / \textbf{3.26} & \cellcolor{ElegantPurple!15}\textbf{73.54} / \textbf{4.34} & \cellcolor{ElegantPurple!15}\textbf{57.14} / \textbf{4.15} & \cellcolor{ElegantPurple!15}\textbf{72.54} / \textbf{3.77} \\
    \bottomrule
  \end{tabularx}
  \endgroup
\end{table}

\begin{table}[t]
  \centering
  \caption{
    \textbf{Simulation results on GR00T in SimplerEnv.}
    Each cell reports success rate (\%, $\uparrow$) / speedup ($\uparrow$).
    \sysname improves the overall success rate from 64.00\% to 65.88\%, with 2.35$\times$ and 2.55$\times$ average speedup on Google Robot and WidowX, respectively.
  }
  \label{tab:groot_main_results}
  \begingroup
  \scriptsize
  \setlength{\tabcolsep}{2pt}
  \renewcommand{\arraystretch}{1.12}
  \begin{tabularx}{\linewidth}{@{}l l *{6}{>{\centering\arraybackslash}X} >{\centering\arraybackslash}X@{}}
    \toprule
    \multirow{2}{*}{\textbf{SIMPLER}} &
    \multirow{2}{*}{\textbf{Method}} &
    \multicolumn{6}{c}{\textbf{Success Rate (\%, $\uparrow$) / Speedup ($\uparrow$)}} &
    \multirow{2}{*}{\textbf{AVG}} \\
    \cmidrule(lr){3-8}
    & & CloseD & MoveNear & OpenD & PickCan & PickObj & PlaceD & \\
    \midrule
    \multirow{4}{*}{\makecell{\textbf{Google}\\\textbf{Robot}}}
    & GR00T~\citep{bjorck2025gr00t} & 82.00 / 1.00 & 92.00 / 1.00 & 49.50 / 1.00 & 99.50 / 1.00 & 92.50 / 1.00 & 11.00 / 1.00 & 71.08 / 1.00 \\
    \cmidrule(lr){2-9}
    & FastV~\citep{chen2024image} & 80.00 / 1.13 & 88.00 / 1.13 & 64.00 / 1.13 & 94.00 / 1.13 & 88.00 / 1.13 & 12.00 / 1.13 & 71.00 / 1.13 \\
    & VLA-Cache~\citep{xu2025vla} & 72.00 / 1.13 & 90.00 / 1.13 & 62.00 / 1.13 & 94.00 / 1.13 & 84.00 / 1.13 & 10.00 / 1.13 & 68.67 / 1.13 \\
    & \cellcolor{ElegantPurple!15}\sysname & \cellcolor{ElegantPurple!15}\textbf{81.00} / \textbf{2.39} & \cellcolor{ElegantPurple!15}\textbf{95.50} / \textbf{1.90} & \cellcolor{ElegantPurple!15}\textbf{74.50} / \textbf{2.31} & \cellcolor{ElegantPurple!15}\textbf{98.00} / \textbf{2.35} & \cellcolor{ElegantPurple!15}\textbf{88.50} / \textbf{2.59} & \cellcolor{ElegantPurple!15}\textbf{12.50} / \textbf{2.39} & \cellcolor{ElegantPurple!15}\textbf{75.00} / \textbf{2.35} \\
    \bottomrule
  \end{tabularx}
  \vspace{2pt}
  \begin{tabularx}{\linewidth}{l l *{7}{>{\centering\arraybackslash}X} >{\centering\arraybackslash}X}
    \toprule
    \multirow{2}{*}{\textbf{SIMPLER}} &
    \multirow{2}{*}{\textbf{Method}} &
    \multicolumn{7}{c}{\textbf{Success Rate (\%, $\uparrow$) / Speedup ($\uparrow$)}} &
    \multirow{2}{*}{\textbf{AVG}} \\
    \cmidrule(lr){3-9}
    & & Carrot & CloseD & OpenD & Basket & Sink & Spoon & Stack & \\
    \midrule
    \multirow{4}{*}{\textbf{WidowX}}
    & GR00T~\citep{bjorck2025gr00t} & 60.50 / 1.00 & 98.50 / 1.00 & 93.00 / 1.00 & 37.50 / 1.00 & 37.50 / 1.00 & 71.50 / 1.00 & 7.00 / 1.00 & 57.93 / 1.00 \\
    \cmidrule(lr){2-10}
    & FastV~\citep{chen2024image} & 58.00 / 1.13 & 98.00 / 1.13 & 84.00 / 1.13 & 22.00 / 1.13 & 30.00 / 1.13 & \textbf{74.00} / 1.13 & 12.00 / 1.13 & 54.00 / 1.13 \\
    & VLA-Cache~\citep{xu2025vla} & \textbf{72.00} / 1.13 & 98.00 / 1.13 & 84.00 / 1.13 & 24.00 / 1.13 & 24.00 / 1.13 & 72.00 / 1.13 & 12.00 / 1.13 & 53.14 / 1.13 \\
    & \cellcolor{ElegantPurple!15}\sysname & \cellcolor{ElegantPurple!15}63.50 / \textbf{2.58} & \cellcolor{ElegantPurple!15}\textbf{99.50} / \textbf{2.54} & \cellcolor{ElegantPurple!15}\textbf{85.00} / \textbf{2.56} & \cellcolor{ElegantPurple!15}\textbf{31.50} / \textbf{2.52} & \cellcolor{ElegantPurple!15}\textbf{41.00} / \textbf{2.57} & \cellcolor{ElegantPurple!15}\textbf{72.50} / \textbf{2.55} & \cellcolor{ElegantPurple!15}\textbf{13.50} / \textbf{2.55} & \cellcolor{ElegantPurple!15}\textbf{58.07} / \textbf{2.55} \\
    \bottomrule
  \end{tabularx}
  \endgroup
\end{table}

\sysname reuses cached refinement updates only on selected steps, allowing the scheduler to reduce action-generation cost without forcing the entire action head to be skipped.
The action-head action space contains three refinement levels, $\ell_t^{\mathrm{H}}\in\{0,1,2\}$.
For each level, we define $\mathcal{R}(\ell_t^{\mathrm{H}})$ as the set of refinement steps whose updates are reused from the cache:
\begin{equation}
    \mathcal{R}(\ell_t^{\mathrm{H}})=
    \begin{cases}
        \varnothing, & \ell_t^{\mathrm{H}}=0,\\
        \{1,\ldots,M-2\}, & \ell_t^{\mathrm{H}}=1,\\
        \{1,\ldots,M-1\}, & \ell_t^{\mathrm{H}}=2,
    \end{cases}
\end{equation}
Thus, $\ell_t^{\mathrm{H}}=0$ recomputes all steps, $\ell_t^{\mathrm{H}}=1$ reuses only middle steps, and $\ell_t^{\mathrm{H}}=2$ reuses all steps after the first refinement.
The scheduled action-head update is
\begin{equation}
    \mathbf{x}_t^{m+1} =
    \begin{cases}
        \mathcal{F}_\phi^m(\mathbf{x}_t^m,\mathbf{z}_t,\mathbf{s}_t), & m \notin \mathcal{R}(\ell_t^{\mathrm{H}}),\\
        \mathbf{x}_t^m+\bar{\boldsymbol{\Delta}}^{m}, & m \in \mathcal{R}(\ell_t^{\mathrm{H}}),
    \end{cases}
\end{equation}
where $\bar{\boldsymbol{\Delta}}^{m}$ is the cached update from a recent execution.

When $\ell_t^{\mathrm{H}}=0$, \sysname recomputes all refinement steps and updates the cache.
When $\ell_t^{\mathrm{H}}=1$, \sysname recomputes the boundary refinement steps and reuses the middle updates, preserving input and output adaptation while reducing intermediate computation.
When $\ell_t^{\mathrm{H}}=2$, \sysname recomputes only the first refinement step and reuses the remaining updates, providing the most aggressive acceleration for smooth action trajectories.
This factorized design enlarges the scheduling space: \sysname can refresh high-level visual-language reasoning while reusing low-level action refinement, or reuse the Vision-LLM backbone while recomputing the action head when precise control is needed.

\subsection{Two-Stage Joint RL Scheduler Design}
\label{sec:rl_scheduler}

\begin{figure}[t]
    \centering
    \includegraphics[width=\linewidth]{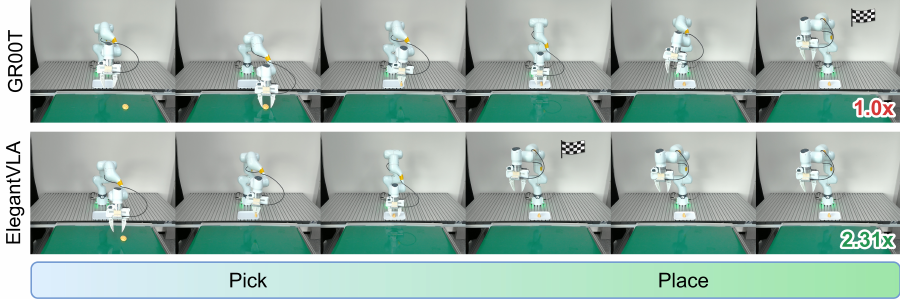}
    \caption{
        \textbf{Qualitative visualization of faster real-world execution.}
        On real-world pineapple-bun pickup, \sysname preserves the pick-and-place phases while completing the task 2.31$\times$ faster than full-computation GR00T.
    }
    \label{fig:real_qualitative_execution}
\end{figure}

\begin{figure}[t]
    \centering
    \includegraphics[width=\linewidth]{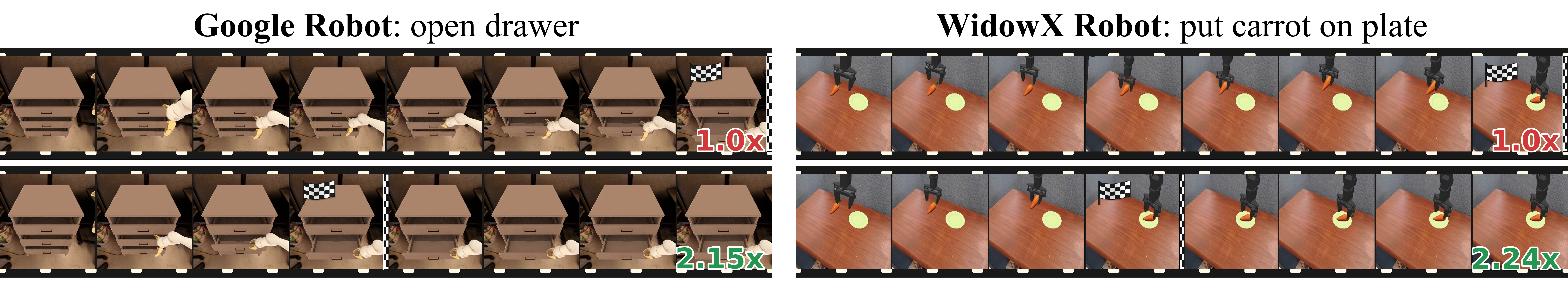}
    \caption{
        \textbf{Qualitative visualization of faster simulated execution.}
        On Google Robot drawer opening and WidowX carrot placement, \sysname completes the rollouts faster while preserving the key manipulation phases.
    }
    \label{fig:sim_qualitative_execution}
\end{figure}

The scheduler serves as the perception-aware compute allocator in \sysname.
Given the current semantic stability and robot motion state, it learns to choose how much to think at each control step, mirroring the human-inspired principle of allocating effort according to movement phase.
This is a sequential decision problem: each action $\mathbf{c}_t=(\ell_t^{\mathrm{B}},\ell_t^{\mathrm{H}})$ jointly selects a Vision-LLM compute level and an action-head refinement level, while aggressive reuse may affect future control accuracy through error accumulation.
The scheduler observation is
\begin{equation}
    \boldsymbol{\xi}_t =
    [\rho_t,\ v_t^{\mathrm{grip}},\ v_t^{\mathrm{trans}},\ v_t^{\mathrm{rot}},\ p_t]^\top,
\end{equation}
where $\rho_t$ is the temporal representation similarity, the next three terms are gripper speed, end-effector translation speed, and end-effector rotation speed, and $p_t=t/T_{\max}$ is the normalized episode progress.
These signals jointly describe high-level semantic stability, low-level motion continuity, and coarse task progress.

We train $\pi_\theta$ with Maskable PPO while keeping the base VLA frozen.
Action masking is important because valid Vision-LLM choices are state-dependent: after a multi-step reuse decision, the backbone level is fixed within the induced reuse window, while the action head still requires a decision at every control step.

Directly optimizing the scheduler from terminal accuracy and FLOPs is difficult because an episode contains hundreds of scheduling decisions but only sparse end-of-episode feedback.
We therefore use a two-stage training procedure.
The first stage uses a rule-guided teacher to provide dense shaping rewards:
\begin{equation}
    r_t^{(1)} =
    r_t^{\mathrm{succ}}
    - \lambda_C r_t^{\mathrm{FLOPs}}
    - \lambda_{\mathrm{B}} d(\ell_t^{\mathrm{B}},\tilde{\ell}_t^{\mathrm{B}})
    - \lambda_{\mathrm{H}} d(\ell_t^{\mathrm{H}},\tilde{\ell}_t^{\mathrm{H}})
    - \lambda_R k(\ell_t^{\mathrm{B}}),
\end{equation}
where $r_t^{\mathrm{FLOPs}}$ measures the computation cost of the selected execution path, $\tilde{\ell}_t^{\mathrm{B}}$ and $\tilde{\ell}_t^{\mathrm{H}}$ are rule-guided computation levels, $d(\cdot,\cdot)$ penalizes disagreement with the teacher when multiple valid choices exist, and $k(\ell_t^{\mathrm{B}})$ denotes the temporal reuse horizon introduced by the chosen Vision-LLM level.
This stage gives the scheduler a stable initialization and prevents destructive exploration.
Stage two removes teacher shaping terms and optimizes task-efficiency feedback, allowing the policy to move beyond hand-designed rules while retaining stable reuse.
At deployment, the teacher and rewards are removed; only the learned scheduler is attached to the frozen VLA policy.

\section{Experiments}
\label{sec:experiments}

\begin{table}[t]
  \centering
  \caption{
    \textbf{Real-world results on Franka Research 3.}
    Each cell reports success rate (\%, $\uparrow$) / speedup ($\uparrow$).
    Results are reported over 10 trials per task with the full-computation GR00T policy as the baseline.
  }
  \label{tab:real_world_results}
  \begingroup
  \scriptsize
  \setlength{\tabcolsep}{2pt}
  \renewcommand{\arraystretch}{1.12}
  \begin{tabularx}{\linewidth}{@{}l l *{6}{>{\centering\arraybackslash}X} >{\centering\arraybackslash}X@{}}
    \toprule
    \multirow{2}{*}{\textbf{Setting}} &
    \multirow{2}{*}{\textbf{Method}} &
    \multicolumn{6}{c}{\textbf{Success Rate (\%, $\uparrow$) / Speedup ($\uparrow$)}} &
    \multirow{2}{*}{\textbf{AVG}} \\
    \cmidrule(lr){3-8}
    & & \makecell{Phone\\Stand} & \makecell{Pen\\Holder} & \makecell{Stack\\Bowls} & \makecell{Pineapple\\Bun} & Toast & Chocolate & \\
    \midrule
    \multirow{2}{*}{\makecell{\textbf{Real}\\\textbf{Robot}}}
    & GR00T~\citep{bjorck2025gr00t} & 60.00 / 1.00 & 70.00 / 1.00 & 40.00 / 1.00 & 70.00 / 1.00 & 70.00 / 1.00 & 60.00 / 1.00 & 61.67 / 1.00 \\
    & \cellcolor{ElegantPurple!15}\sysname & \cellcolor{ElegantPurple!15}60.00 / \textbf{2.18} & \cellcolor{ElegantPurple!15}70.00 / \textbf{2.24} & \cellcolor{ElegantPurple!15}40.00 / \textbf{2.01} & \cellcolor{ElegantPurple!15}\textbf{80.00} / \textbf{2.31} & \cellcolor{ElegantPurple!15}\textbf{80.00} / \textbf{2.21} & \cellcolor{ElegantPurple!15}60.00 / \textbf{2.11} & \cellcolor{ElegantPurple!15}\textbf{65.00} / \textbf{2.18} \\
    \bottomrule
  \end{tabularx}
  \endgroup
\end{table}

\begin{table}[t]
  \centering
  \caption{
    \textbf{Latency and frequency on GR00T in SimplerEnv.}
    Each cell reports frequency (Hz, $\uparrow$) / latency (ms, $\downarrow$), measured on an RTX 4090.
    \sysname more than doubles the average control frequency while reducing per-step latency by over half on both robot suites.
  }
  \label{tab:latency_frequency_simplerenv}
  \begingroup
  \scriptsize
  \setlength{\tabcolsep}{2pt}
  \renewcommand{\arraystretch}{1.08}
  \begin{tabularx}{\linewidth}{@{}l l *{6}{>{\centering\arraybackslash}X} >{\centering\arraybackslash}X@{}}
    \toprule
    \multirow{2}{*}{\textbf{SIMPLER}} &
    \multirow{2}{*}{\textbf{Method}} &
    \multicolumn{6}{c}{\textbf{Frequency (Hz, $\uparrow$) / Latency (ms, $\downarrow$)}} &
    \multirow{2}{*}{\textbf{AVG}} \\
    \cmidrule(lr){3-8}
    & & CloseD & MoveNear & OpenD & PickCan & PickObj & PlaceD & \\
    \midrule
    \multirow{2}{*}{\makecell{\textbf{Google}\\\textbf{Robot}}}
    & GR00T~\citep{bjorck2025gr00t} & 15.99 / 62.56 & 16.17 / 61.83 & 17.62 / 56.75 & 15.88 / 62.96 & 16.90 / 59.18 & 17.46 / 57.28 & 16.64 / 60.09 \\
    & \cellcolor{ElegantPurple!15}\sysname & \cellcolor{ElegantPurple!15}\textbf{36.36} / \textbf{27.50} & \cellcolor{ElegantPurple!15}\textbf{27.19} / \textbf{36.78} & \cellcolor{ElegantPurple!15}\textbf{35.86} / \textbf{27.88} & \cellcolor{ElegantPurple!15}\textbf{36.20} / \textbf{27.63} & \cellcolor{ElegantPurple!15}\textbf{38.20} / \textbf{26.18} & \cellcolor{ElegantPurple!15}\textbf{36.36} / \textbf{27.50} & \cellcolor{ElegantPurple!15}\textbf{35.03} / \textbf{28.55} \\
    \bottomrule
  \end{tabularx}
  \par\vspace{3pt}
  {
  \fontsize{6.0}{6.8}\selectfont
  \setlength{\tabcolsep}{1pt}
  \renewcommand{\arraystretch}{1.02}
  \noindent\resizebox{\linewidth}{!}{%
  \begin{tabular}{@{}l l *{7}{c} c@{}}
    \toprule
    \multirow{2}{*}{\textbf{SIMPLER}} &
    \multirow{2}{*}{\textbf{Method}} &
    \multicolumn{7}{c}{\textbf{Frequency (Hz, $\uparrow$) / Latency (ms, $\downarrow$)}} &
    \multirow{2}{*}{\textbf{AVG}} \\
    \cmidrule(lr){3-9}
    & & Carrot & CloseD & OpenD & Basket & Sink & Spoon & Stack & \\
    \midrule
    \multirow{2}{*}{\textbf{WidowX}}
    & GR00T~\citep{bjorck2025gr00t} & 17.26 / 57.94 & 17.25 / 57.98 & 17.18 / 58.21 & 15.72 / 63.60 & 17.07 / 58.57 & 17.48 / 57.21 & 15.16 / 65.96 & 16.69 / 59.92 \\
    & \cellcolor{ElegantPurple!15}\sysname & \cellcolor{ElegantPurple!15}\textbf{37.36} / \textbf{26.76} & \cellcolor{ElegantPurple!15}\textbf{37.86} / \textbf{26.41} & \cellcolor{ElegantPurple!15}\textbf{37.20} / \textbf{26.88} & \cellcolor{ElegantPurple!15}\textbf{37.70} / \textbf{26.53} & \cellcolor{ElegantPurple!15}\textbf{38.20} / \textbf{26.18} & \cellcolor{ElegantPurple!15}\textbf{37.03} / \textbf{27.00} & \cellcolor{ElegantPurple!15}\textbf{38.03} / \textbf{26.29} & \cellcolor{ElegantPurple!15}\textbf{37.70} / \textbf{26.53} \\
    \bottomrule
  \end{tabular}%
  }%
  \par
  }
  \endgroup
\end{table}

\subsection{Main Results}
\label{sec:main_results}

\textbf{Simulation results.} 
Table~\ref{tab:cogact_main_results} shows that \sysname achieves the best average success-speedup trade-off on CogACT.
On Visual Matching and Variant Aggregation, \sysname reaches 77.59\% and 72.54\% average success with 3.72$\times$ and 3.77$\times$ speedup, respectively, outperforming existing acceleration baselines that provide only 1.20$\times$--1.53$\times$ speedup.
The gains are especially clear on drawer-style tasks, where adaptive recomputation helps preserve control accuracy during contact-sensitive phases.
Table~\ref{tab:groot_main_results} further verifies the same trend on GR00T, where \sysname improves the overall success rate from 64.00\% to 65.88\% while achieving up to 2.55$\times$ average FLOPs speedup; Appendix~\ref{app:simulation_setup} gives the simulation task protocol.

\textbf{Real-world results.} We evaluate \sysname on six real-robot tasks that span both stationary manipulation and conveyor-belt pickup: placing a phone on a stand, inserting a pen into a holder, stacking three bowls, and transferring food items from a moving conveyor belt to a plate in front of the robot arm.
Table~\ref{tab:real_world_results} shows that \sysname transfers well to the physical setting and preserves the efficiency gains observed in simulation.
Overall, it improves the average success rate from 61.67\% to 65.00\% while achieving a 2.18$\times$ speedup.
The gain is most apparent on conveyor-belt pickup tasks, suggesting that faster control helps \sysname react to moving targets; Appendix~\ref{app:real_world} gives detailed real-world task descriptions, success rates, and acceleration results.

\subsection{Analysis}
\label{sec:analysis}

\textbf{Frequency and Latency.}
Table~\ref{tab:latency_frequency_simplerenv} shows that the computation reduction of \sysname translates into practical wall-clock gains.
On Google Robot, \sysname increases the average control frequency from 16.64 Hz to 35.03 Hz and reduces latency from 60.09 ms to 28.55 ms.
On WidowX, it improves the average frequency from 16.69 Hz to 37.70 Hz and reduces latency from 59.92 ms to 26.53 ms.
The gains are consistent across tasks, while the lower frequency on MoveNear suggests that the scheduler remains conservative when continuous spatial reasoning is needed.

\begin{table}[t]
  \centering
  \caption{
    \textbf{RL training ablation on CogACT in SimplerEnv.}
    Each cell reports success rate (\%, $\uparrow$) / speedup ($\uparrow$).
    Stage2 improves the average success rate from 70.45\% to 77.59\% on Visual Matching and from 69.03\% to 72.54\% on Visual Aggregation while also increasing the average speedup.
  }
  \label{tab:rl_ablation_cogact}
  \begingroup
  \footnotesize
  \setlength{\tabcolsep}{2pt}
  \renewcommand{\arraystretch}{1.08}
  \begin{tabularx}{\linewidth}{l l *{5}{>{\centering\arraybackslash}X}}
    \toprule
    \multirow{2}{*}{\textbf{SIMPLER}} &
    \multirow{2}{*}{\textbf{Method}} &
    \multicolumn{4}{c}{\textbf{Success Rate (\%, $\uparrow$) / Speedup ($\uparrow$)}} &
    \multirow{2}{*}{\textbf{AVG}} \\
    \cmidrule(lr){3-6}
    & & PickCan & MoveNear & Drawer & PutDrawer & \\
    \midrule
    \multirow{3}{*}{\makecell{\textbf{Visual}\\\textbf{Matching}}}
    & CogACT~\citep{li2024cogact} & 94.33 / 1.00 & 83.75 / 1.00 & 74.04 / 1.00 & 40.74 / 1.00 & 73.22 / 1.00 \\
    \cmidrule(lr){2-7}
    & Stage1 & 88.00 / 3.07 & 72.50 / \textbf{3.12} & 88.89 / 3.10 & 32.41 / 4.40 & 70.45 / 3.42 \\
    & \cellcolor{ElegantPurple!15}Stage2 & \cellcolor{ElegantPurple!15}\textbf{89.67} / \textbf{3.32} & \cellcolor{ElegantPurple!15}\textbf{80.42} / 3.05 & \cellcolor{ElegantPurple!15}\textbf{89.35} / \textbf{4.10} & \cellcolor{ElegantPurple!15}\textbf{50.93} / \textbf{4.41} & \cellcolor{ElegantPurple!15}\textbf{77.59} / \textbf{3.72} \\
    \midrule
    \multirow{3}{*}{\makecell{\textbf{Visual}\\\textbf{Aggregation}}}
    & CogACT~\citep{li2024cogact} & 88.97 / 1.00 & 77.00 / 1.00 & 27.51 / 1.00 & 55.03 / 1.00 & 62.13 / 1.00 \\
    \cmidrule(lr){2-7}
    & Stage1 & 81.09 / 3.08 & 70.67 / 3.12 & 70.90 / 3.12 & 53.44 / 3.90 & 69.03 / 3.31 \\
    & \cellcolor{ElegantPurple!15}Stage2 & \cellcolor{ElegantPurple!15}\textbf{82.30} / \textbf{3.32} & \cellcolor{ElegantPurple!15}\textbf{77.17} / \textbf{3.26} & \cellcolor{ElegantPurple!15}\textbf{73.54} / \textbf{4.34} & \cellcolor{ElegantPurple!15}\textbf{57.14} / \textbf{4.15} & \cellcolor{ElegantPurple!15}\textbf{72.54} / \textbf{3.77} \\
    \bottomrule
  \end{tabularx}
  \endgroup
\end{table}

\textbf{RL training.} Table~\ref{tab:rl_ablation_cogact} validates the role of the two-stage scheduler training.
Compared with Stage1, Stage2 improves the average success rate from 70.45\% to 77.59\% on Visual Matching and from 69.03\% to 72.54\% on Visual Aggregation, while also increasing the average speedup.
This indicates that RL learns a better task-efficiency trade-off after rule-guided initialization, rather than simply applying more aggressive reuse; Appendix~\ref{app:scheduler_training_ablation} ablates threshold-only scheduling, Stage1 training, and two-stage training, further highlighting the effectiveness of \sysname.

\begin{table}[t]
  \centering
  \caption{
    \textbf{Module scheduling ablation on GR00T Google Robot tasks.}
    Each cell reports success rate (\%, $\uparrow$) / speedup ($\uparrow$).
    Jointly scheduling both branches achieves the best average success and substantially larger speedup than keeping either branch at full computation.
  }
  \label{tab:module_ablation_groot}
  \begingroup
  \scriptsize
  \setlength{\tabcolsep}{2pt}
  \renewcommand{\arraystretch}{1.08}
  \begin{tabularx}{\linewidth}{l l *{6}{>{\centering\arraybackslash}X} >{\centering\arraybackslash}X}
    \toprule
    \multirow{2}{*}{\textbf{SIMPLER}} &
    \multirow{2}{*}{\textbf{Method}} &
    \multicolumn{6}{c}{\textbf{Success Rate (\%, $\uparrow$) / Speedup ($\uparrow$)}} &
    \multirow{2}{*}{\textbf{AVG}} \\
    \cmidrule(lr){3-8}
    & & CloseD & MoveNear & OpenD & PickCan & PickObj & PlaceD & \\
    \midrule
    \multirow{4}{*}{\makecell{\textbf{Google}\\\textbf{Robot}}}
    & GR00T~\citep{bjorck2025gr00t} & 82.00 / 1.00 & 92.00 / 1.00 & 49.50 / 1.00 & 99.50 / 1.00 & 92.50 / 1.00 & 11.00 / 1.00 & 71.08 / 1.00 \\
    \cmidrule(lr){2-9}
    & Force LLM Full & 73.50 / 1.47 & 90.00 / 1.47 & 47.50 / 1.47 & 96.00 / 1.46 & 88.00 / 1.47 & \textbf{14.00} / 1.45 & 68.17 / 1.46 \\
    & Force AH Full & \textbf{86.00} / 1.56 & 87.00 / 1.56 & 70.50 / 1.56 & \textbf{98.00} / 1.55 & 87.00 / 1.56 & 8.00 / 1.56 & 72.75 / 1.56 \\
    & \cellcolor{ElegantPurple!15}\sysname & \cellcolor{ElegantPurple!15}81.00 / \textbf{2.39} & \cellcolor{ElegantPurple!15}\textbf{95.50} / \textbf{1.90} & \cellcolor{ElegantPurple!15}\textbf{74.50} / \textbf{2.31} & \cellcolor{ElegantPurple!15}\textbf{98.00} / \textbf{2.35} & \cellcolor{ElegantPurple!15}\textbf{88.50} / \textbf{2.59} & \cellcolor{ElegantPurple!15}12.50 / \textbf{2.39} & \cellcolor{ElegantPurple!15}\textbf{75.00} / \textbf{2.35} \\
    \bottomrule
  \end{tabularx}
  \endgroup
\end{table}

\textbf{Module scheduling.} Table~\ref{tab:module_ablation_groot} studies whether the Vision-LLM backbone and action head should be scheduled jointly.
Keeping either branch at full computation limits the benefit: Force LLM Full obtains 68.17\% success with 1.46$\times$ speedup, and Force AH Full obtains 72.75\% success with 1.56$\times$ speedup.
The full \sysname reaches 75.00\% success with 2.35$\times$ speedup, showing that semantic reasoning and action refinement contain complementary temporal redundancy; Appendix~\ref{app:scheduler_observation_ablation} ablates CKA- and speed-based scheduler inputs, confirming their complementarity.

\textbf{Qualitative execution behavior.}
Figures~\ref{fig:real_qualitative_execution} and~\ref{fig:sim_qualitative_execution} show that \sysname finishes representative rollouts faster while preserving key manipulation phases.
The visualized speedups, 2.31$\times$ on real pineapple-bun pickup and about 2.2$\times$ in simulation, suggest that the scheduler removes redundant inference during stable motion without skipping precision-sensitive transitions such as approach, grasp/contact, and placement.
Additional rollouts are provided in Appendices~\ref{app:real_world_rollout_visualizations} and~\ref{app:simulation_rollout_visualizations}.

\section{Conclusion}
\label{sec:conclusion}

We presented \sysname, a phase-adaptive inference framework that learns when to recompute and when to reuse computation in frozen VLA policies.
By jointly scheduling the Vision-LLM backbone and the action head, \sysname allocates more computation to precision-sensitive phases and avoids redundant inference during stable motion.
Across CogACT, GR00T, and real-world manipulation, \sysname achieves substantial speedups, up to 3.77$\times$ in simulation and 2.18$\times$ on the real robot, while preserving competitive task success.
These results suggest that efficient embodied policies should not only reduce per-step cost, but also adapt computation to the temporal demands of interaction.

\newpage
\bibliographystyle{unsrtnat}
\bibliography{reference}

\newpage
\appendix
\section*{Limitations and Responsible Use}

This work is a technical study of inference-time acceleration for already trained vision-language-action robot policies.
It does not involve new human-subject data, personal data, or deployment decisions, and we do not identify direct social or ethical concerns specific to the proposed scheduling method.
As an acceleration method, \sysname aims to reduce redundant computation while preserving the behavior of the underlying policy; real-world use should still follow the same task validation, robot safety checks, and fallback execution required for the base policy.

\section{Real-World Protocol and Additional Results}
\label{app:real_world}

\begin{figure}[htbp]
    \centering
    \includegraphics[width=\linewidth]{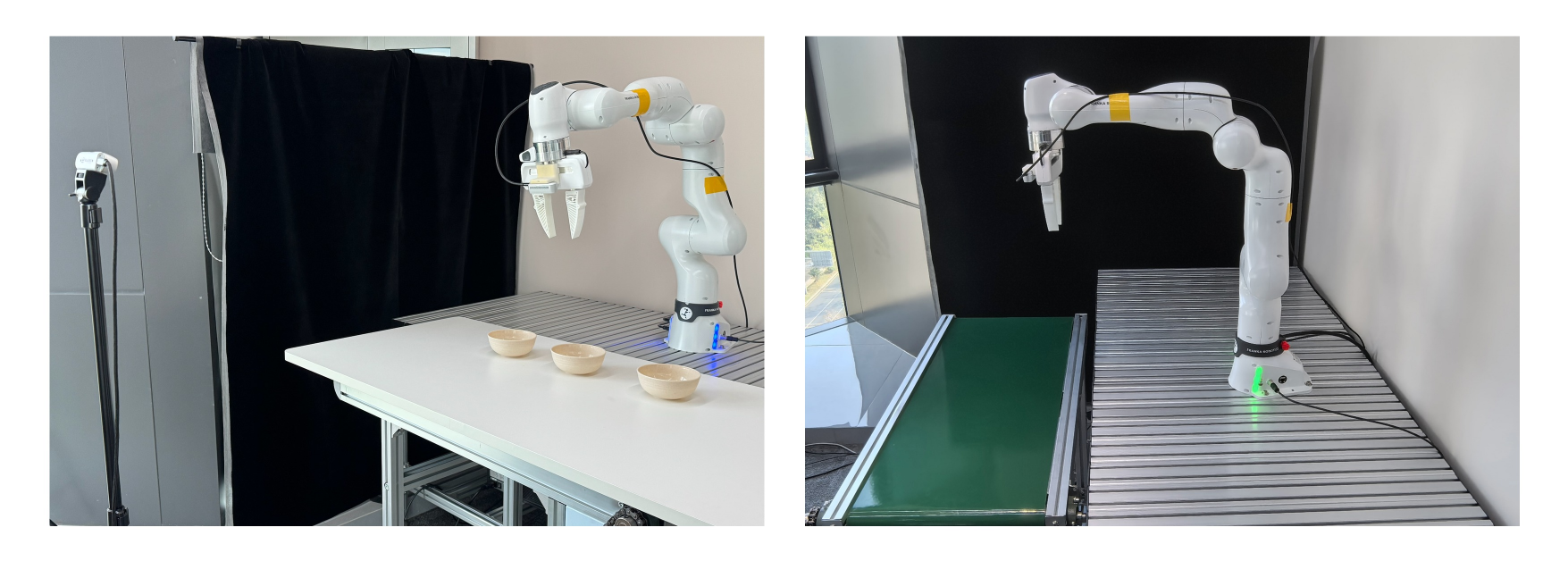}
    \caption{
        \textbf{Real-world experimental platform.}
        The physical experiments use a Franka Research 3 setup with stationary tabletop manipulation on the left and conveyor-belt pickup on the right.
    }
    \label{fig:app_real_robot_platform}
\end{figure}

\begin{table}[htbp]
  \centering
    \caption{
        \textbf{Real-world task prompts and success criteria.}
        Natural-language prompts and success criteria used for the six tasks in Table~\ref{tab:real_world_results}.
  }
  \label{tab:app_real_task_definitions}
  \begingroup
  \scriptsize
  \setlength{\tabcolsep}{3pt}
  \renewcommand{\arraystretch}{1.12}
  \begin{tabularx}{\linewidth}{>{\raggedright\arraybackslash}p{0.16\linewidth} >{\raggedright\arraybackslash}p{0.34\linewidth} >{\raggedright\arraybackslash}p{0.42\linewidth}}
    \toprule
    \textbf{Task} & \textbf{Instruction prompt} & \textbf{Success criterion} \\
    \midrule
    Phone Stand & Place the phone on a stand & The phone-like object is placed on the stand and remains stable. \\
    Pen Holder & Insert the pen into the pen holder & The pen-like object is inserted into the holder without displacing it into a failed configuration. \\
    Stack Bowls & Stack the three bowls on the table together & The three bowls are stacked together and remain stable after release. \\
    Pineapple Bun & move the pineapple bun from the conveyor belt to the center of the plate & The pineapple bun is moved from the conveyor belt to the center of the plate. \\
    Toast & move the toast from the conveyor belt to the center of the plate & The toast is moved from the conveyor belt to the center of the plate. \\
    Chocolate & move the chocolate bun from the conveyor belt to the center of the plate & The chocolate bun is moved from the conveyor belt to the center of the plate. \\
    \bottomrule
  \end{tabularx}
  \endgroup
\end{table}

\paragraph{Platform and protocol.}
Fig.~\ref{fig:app_real_robot_platform} shows the physical setup.
All real-world experiments are conducted on a Franka Research 3 robot equipped with Intel RealSense D435i cameras.
The robot is controlled through Polymetis, while the GR00T-based policy server runs high-level VLA inference on an NVIDIA GeForce RTX 4090 GPU.
We use both a front-mounted eye-to-hand camera and a wrist-mounted eye-in-hand camera.
For each task and each method, we run 10 trials under the same initialization protocol.
The full-computation GR00T policy is used as the real-world baseline.
A trial is counted as successful only when the final object state satisfies the task goal and remains stable after execution.

\paragraph{Task design and analysis.}
Table~\ref{tab:app_real_task_definitions} lists the instruction prompts and success criteria for the six real-world tasks.
These tasks separate the evaluation into two regimes aligned with our phase-adaptive view.
Phone Stand, Pen Holder, and Stack Bowls are stationary precision tasks, testing whether \sysname preserves accurate contact, placement, and release while removing redundant computation.
Pineapple Bun, Toast, and Chocolate Bun are conveyor-belt pickup tasks, where moving targets make delayed actions costly and faster inference should improve closed-loop correction.
Together, they test whether stable phases can tolerate temporal reuse while precision- or correction-sensitive phases still receive fresh reasoning and action refinement.

\paragraph{Additional real-world results.}
Tables~\ref{tab:app_real_success_counts} and~\ref{tab:app_real_latency} decompose the real-world result in Table~\ref{tab:real_world_results} into trial counts and latency measurements.
Table~\ref{tab:app_real_success_counts} shows that \sysname matches full-computation GR00T on the three stationary precision tasks, indicating that adaptive reuse does not noticeably degrade contact-sensitive manipulation in this setting.
The only success-count improvements appear on Pineapple Bun and Toast, where the moving target makes lower-latency feedback more valuable.
Table~\ref{tab:app_real_latency} confirms that these gains are backed by practical runtime reductions: average per-step latency drops from 72.44 ms to 38.00 ms, corresponding to a 1.91$\times$ wall-clock speedup and 2.18$\times$ FLOPs speedup.
This pattern is consistent with the main analysis: \sysname converts redundant computation into faster control while keeping enough fresh reasoning and action refinement for precision-sensitive phases.

\begin{table}[t]
  \centering
  \caption{
    \textbf{Trial-level breakdown for the main real-world results.}
    Each task is evaluated over 10 real-robot trials, and each cell reports successful trials.
    The improvement row reports the change relative to full-computation GR00T, showing that \sysname preserves success on stationary precision tasks while improving conveyor-belt pickup.
  }
  \label{tab:app_real_success_counts}
  \begingroup
  \small
  \setlength{\tabcolsep}{3pt}
  \renewcommand{\arraystretch}{1.10}
  \begin{tabularx}{\linewidth}{l *{8}{>{\centering\arraybackslash}X}}
    \toprule
    \textbf{Method} & \textbf{Phone} & \textbf{Pen} & \textbf{Bowls} & \makecell{\textbf{Pine}\\\textbf{apple}} & \textbf{Toast} & \makecell{\textbf{Choco}\\\textbf{late}} & \textbf{Total} & \textbf{Rate} \\
    \midrule
    Full computation & 6/10 & 7/10 & 4/10 & 7/10 & 7/10 & 6/10 & 37/60 & 61.67\% \\
    \rowcolor{ElegantPurple!15}\sysname & 6/10 & 7/10 & 4/10 & 8/10 & 8/10 & 6/10 & 39/60 & \textbf{65.00\%} \\
    Improvement & +0 & +0 & +0 & +1 & +1 & +0 & +2 & +3.33 pp \\
    \bottomrule
  \end{tabularx}
  \endgroup
\end{table}

\begin{table}[t]
  \centering
  \caption{
    \textbf{Latency breakdown for the main real-world results.}
    We report per-step wall-clock latency for full-computation GR00T and \sysname, the corresponding latency reduction, wall-clock speedup, and FLOPs speedup.
    The results show that phase-adaptive reuse consistently reduces real-robot inference latency, with the largest gains aligning with tasks that benefit from faster closed-loop correction.
  }
  \label{tab:app_real_latency}
  \begingroup
  \small
  \setlength{\tabcolsep}{4pt}
  \renewcommand{\arraystretch}{1.12}
  \begin{tabularx}{\linewidth}{l *{6}{>{\centering\arraybackslash}X}}
    \toprule
    \textbf{Task} & \textbf{Success} & \textbf{Full GR00T} & \textbf{\sysname} & \textbf{Reduction} & \textbf{Latency} & \textbf{FLOPs} \\
    & \textbf{rate} & \textbf{ms/step} & \textbf{ms/step} & & \textbf{speedup} & \textbf{speedup} \\
    \midrule
    Phone-on-Stand & 60.0\% & 72.31 & 37.70 & 47.9\% & 1.92$\times$ & 2.18$\times$ \\
    Pen-to-Holder & 70.0\% & 71.74 & 36.20 & 49.5\% & 1.98$\times$ & 2.24$\times$ \\
    Stack-Bowls & 40.0\% & 71.92 & 41.30 & 42.6\% & 1.74$\times$ & 2.01$\times$ \\
    Pineapple-Bun & 80.0\% & 72.74 & 35.70 & 50.9\% & 2.04$\times$ & 2.31$\times$ \\
    Toast & 80.0\% & 72.88 & 37.40 & 48.7\% & 1.95$\times$ & 2.21$\times$ \\
    Chocolate-Bun & 60.0\% & 73.05 & 39.70 & 45.7\% & 1.84$\times$ & 2.11$\times$ \\
    \midrule
    Average & 65.0\% & 72.44 & 38.00 & 47.5\% & 1.91$\times$ & 2.18$\times$ \\
    \bottomrule
  \end{tabularx}
  \endgroup
\end{table}

\clearpage
\section{Simulation Environment and Task Protocol}
\label{app:simulation_setup}

\begin{table}[t]
  \centering
  \caption{
    \textbf{CogACT simulation task groups.}
    We list the task headers used in the CogACT VM and VA result tables.
    The VM episode counts for PickCan, MoveNear, Drawer, and PutDrawer are 300/240/216/108, and the corresponding VA episode counts are 825/600/189/189.
  }
  \label{tab:app_cogact_task_abbrev}
  \begingroup
  \small
  \setlength{\tabcolsep}{4pt}
  \renewcommand{\arraystretch}{1.15}
  \begin{tabularx}{\linewidth}{p{0.16\linewidth} p{0.32\linewidth} p{0.43\linewidth}}
    \toprule
    \textbf{Header} & \textbf{Expanded task group} & \textbf{Meaning in evaluation} \\
    \midrule
    PickCan & Pick Coke Can & The Google Robot grasps an opened Coke can under object-pose, distractor, background, lighting, and camera variants. \\
    MoveNear & Move Near & The Google Robot moves a target object near a reference target, testing spatial relational reasoning and continuous closed-loop correction. \\
    Drawer & Drawer Manipulation & Aggregates drawer opening and closing tasks over top, middle, and bottom drawer variants. \\
    PutDrawer & Put in Drawer & The Google Robot places an object, such as an apple, into a closed drawer; this task combines grasping, drawer geometry, and final placement. \\
    \bottomrule
  \end{tabularx}
  \endgroup
\end{table}

\begin{table}[t]
  \centering
  \caption{
    \textbf{GR00T simulation task abbreviations by robot suite.}
    Headers are used in the Google Robot and WidowX Robot result tables, with 200 simulation episodes evaluated for each task.
  }
  \label{tab:app_groot_task_abbrev}
  \label{tab:app_groot_google_task_abbrev}
  \label{tab:app_groot_widowx_task_abbrev}
  \begingroup
  \small
  \setlength{\tabcolsep}{3pt}
  \renewcommand{\arraystretch}{1.12}
  \begin{tabularx}{\linewidth}{>{\raggedright\arraybackslash}p{0.18\linewidth} >{\raggedright\arraybackslash}p{0.12\linewidth} >{\raggedright\arraybackslash}p{0.24\linewidth} >{\raggedright\arraybackslash}X}
    \toprule
    \textbf{Robot} & \textbf{Header} & \textbf{Expanded task name} & \textbf{Task meaning} \\
    \midrule
    Google Robot & CloseD & Close Drawer & Close the drawer to the target final state. \\
    Google Robot & MoveNear & Move Near & Move an object near a specified target or reference region. \\
    Google Robot & OpenD & Open Drawer & Open the drawer to the required extent. \\
    Google Robot & PickCan & Pick Coke Can & Pick or grasp a Coke can from the tabletop. \\
    Google Robot & PickObj & Pick Object & Pick a generic target object rather than a fixed Coke-can object. \\
    Google Robot & PlaceD & Place in Closed Drawer & Place the target object into a closed drawer. \\
    \midrule
    WidowX Robot & Carrot & Carrot on Plate & Place the carrot on the plate. \\
    WidowX Robot & CloseD & Close Drawer & Close the drawer in the WidowX setup. \\
    WidowX Robot & OpenD & Open Drawer & Open the drawer in the WidowX setup. \\
    WidowX Robot & Basket & Eggplant in Basket & Put the eggplant into the basket. \\
    WidowX Robot & Sink & Eggplant in Sink & Put the eggplant into the sink. \\
    WidowX Robot & Spoon & Spoon on Towel & Place the spoon on the towel. \\
    WidowX Robot & Stack & Stack Cube & Stack one cube on another cube. \\
    \bottomrule
  \end{tabularx}
  \endgroup
\end{table}

\paragraph{Benchmark protocol.}
All simulation experiments follow the SimplerEnv grouping used in the main tables.
We evaluate frozen CogACT and GR00T backbones against their full-computation baselines, changing only the inference execution path.
Profiling is performed on an NVIDIA RTX 4090 GPU.

\paragraph{Task groups.}
GR00T covers six Google Robot tasks and seven WidowX Robot tasks, with 200 episodes per task.
CogACT uses the VM and VA task groups: PickCan, MoveNear, Drawer, and PutDrawer, with episode counts 300/240/216/108 for VM and 825/600/189/189 for VA.
Tables~\ref{tab:app_cogact_task_abbrev} and~\ref{tab:app_groot_task_abbrev} expand the compact task headers used in the main and appendix simulation tables.
Together, these tasks cover contact-sensitive manipulation and smoother transport phases, matching the phase-adaptive evaluation goal of \sysname.

\clearpage
\section{Additional Scheduler Ablations}
\label{app:scheduler_ablations}

This section supplements the main ablation study with two scheduler-specific questions.
First, we examine whether a hand-designed threshold rule or only the first RL training stage is sufficient.
Second, we ablate the scheduler observations to test why CKA, robot speed, and task progress are used together.
All GR00T results are evaluated in SimplerEnv, and each cell reports success rate and FLOPs speedup.

\subsection{Effect of Rule-Based and Two-Stage RL Scheduling}
\label{app:scheduler_training_ablation}

Table~\ref{tab:app_scheduler_training_ablation} compares the full-computation baseline, a rule-based threshold scheduler, the Stage1 RL scheduler, and the final two-stage \sysname scheduler.
The threshold scheduler tests whether a manually specified representation-similarity trigger is sufficient.
Stage1 RL tests whether the first training stage gives a usable initialization, while \sysname further applies the task-feedback correction from Stage2.

\begin{table}[htpb]
  \centering
  \caption{
    \textbf{Effect of rule-based and two-stage RL scheduling on GR00T.}
    Each cell reports success rate (\%, $\uparrow$) / FLOPs speedup ($\uparrow$).
    Threshold is a hand-designed scheduler, Stage1 RL is the first scheduler-training stage, and \sysname is the complete two-stage scheduler.
  }
  \label{tab:app_scheduler_training_ablation}
  \begingroup
  \scriptsize
  \setlength{\tabcolsep}{1pt}
  \renewcommand{\arraystretch}{1.08}
  \begin{tabularx}{\linewidth}{@{}l l *{6}{>{\centering\arraybackslash}X} >{\centering\arraybackslash}X@{}}
    \toprule
    \multirow{2}{*}{\textbf{SIMPLER}} &
    \multirow{2}{*}{\textbf{Method}} &
    \multicolumn{6}{c}{\textbf{Success Rate (\%, $\uparrow$) / FLOPs Speedup ($\uparrow$)}} &
    \multirow{2}{*}{\textbf{AVG}} \\
    \cmidrule(lr){3-8}
    & & CloseD & MoveNear & OpenD & PickCan & PickObj & PlaceD & \\
    \midrule
    \multirow{4}{*}{\makecell{\textbf{Google}\\\textbf{Robot}}}
    & Baseline & 82.00 / 1.00 & 92.00 / 1.00 & 49.50 / 1.00 & 99.50 / 1.00 & 92.50 / 1.00 & 11.00 / 1.00 & 71.08 / 1.00 \\
    \cmidrule(lr){2-9}
    & Threshold & 80.50 / 2.08 & 91.50 / 1.97 & 67.50 / 2.12 & 97.00 / 1.91 & 87.00 / 2.03 & 12.00 / 2.14 & 72.58 / 2.05 \\
    & Stage1 RL & 80.50 / 1.98 & 80.00 / \textbf{2.43} & 72.50 / 2.03 & \textbf{98.50} / 1.85 & \textbf{90.00} / 1.99 & \textbf{15.00} / 2.12 & 72.75 / 2.09 \\
    & \cellcolor{ElegantPurple!15}\sysname & \cellcolor{ElegantPurple!15}\textbf{81.00} / \textbf{2.39} & \cellcolor{ElegantPurple!15}\textbf{95.50} / 1.90 & \cellcolor{ElegantPurple!15}\textbf{74.50} / \textbf{2.31} & \cellcolor{ElegantPurple!15}98.00 / \textbf{2.35} & \cellcolor{ElegantPurple!15}88.50 / \textbf{2.59} & \cellcolor{ElegantPurple!15}12.50 / \textbf{2.39} & \cellcolor{ElegantPurple!15}\textbf{75.00} / \textbf{2.35} \\
    \bottomrule
  \end{tabularx}
  \vspace{2pt}
  \begin{tabularx}{\linewidth}{@{}l l *{7}{>{\centering\arraybackslash}X} >{\centering\arraybackslash}X@{}}
    \toprule
    \multirow{2}{*}{\textbf{SIMPLER}} &
    \multirow{2}{*}{\textbf{Method}} &
    \multicolumn{7}{c}{\textbf{Success Rate (\%, $\uparrow$) / FLOPs Speedup ($\uparrow$)}} &
    \multirow{2}{*}{\textbf{AVG}} \\
    \cmidrule(lr){3-9}
    & & Carrot & CloseD & OpenD & Basket & Sink & Spoon & Stack & \\
    \midrule
    \multirow{4}{*}{\textbf{WidowX}}
    & Baseline & 60.50 / 1.00 & 98.50 / 1.00 & 93.00 / 1.00 & 37.50 / 1.00 & 37.50 / 1.00 & 71.50 / 1.00 & 7.00 / 1.00 & 57.93 / 1.00 \\
    \cmidrule(lr){2-10}
    & Threshold & 62.00 / 2.13 & 99.00 / 2.13 & 84.50 / 2.20 & 27.50 / 1.97 & 25.00 / 2.24 & 72.00 / 2.18 & 13.00 / 2.23 & 54.71 / 2.12 \\
    & Stage1 RL & 63.00 / \textbf{2.59} & 99.00 / 2.52 & 83.50 / \textbf{2.57} & 15.00 / \textbf{2.53} & 40.00 / \textbf{2.57} & 67.00 / \textbf{2.55} & 11.50 / 2.54 & 54.14 / \textbf{2.55} \\
    & \cellcolor{ElegantPurple!15}\sysname & \cellcolor{ElegantPurple!15}\textbf{63.50} / 2.58 & \cellcolor{ElegantPurple!15}\textbf{99.50} / \textbf{2.54} & \cellcolor{ElegantPurple!15}\textbf{85.00} / 2.56 & \cellcolor{ElegantPurple!15}\textbf{31.50} / 2.52 & \cellcolor{ElegantPurple!15}\textbf{41.00} / \textbf{2.57} & \cellcolor{ElegantPurple!15}\textbf{72.50} / \textbf{2.55} & \cellcolor{ElegantPurple!15}\textbf{13.50} / \textbf{2.55} & \cellcolor{ElegantPurple!15}\textbf{58.07} / \textbf{2.55} \\
    \bottomrule
  \end{tabularx}
  \endgroup
\end{table}

\paragraph{Analysis.}
The threshold scheduler confirms that a representation-similarity trigger is useful, but it is not sufficient as a final policy.
It slightly improves the Google Robot average from 71.08\% to 72.58\%, yet drops WidowX from 57.93\% to 54.71\%, showing that a fixed hand rule does not transfer cleanly across manipulation regimes.
Stage1 RL gives the scheduler a usable initialization and often increases acceleration; on WidowX, it reaches a 2.55$\times$ average speedup.
However, Stage1 can still be too aggressive: on Google MoveNear, it increases speedup to 2.43$\times$ but reduces success to 80.00\%.
The complete two-stage scheduler recovers the trade-off by using Stage2 task feedback to calibrate reuse, improving Google Robot to 75.00\% and WidowX to 58.07\%.
Thus, the final gain is not simply from reusing more computation, but from learning when reuse should be relaxed to preserve task success.

\subsection{Effect of Scheduler Observations}
\label{app:scheduler_observation_ablation}

Table~\ref{tab:app_scheduler_observation_ablation} ablates the scheduler input signals.
Episode progress is included in all reduced-observation variants as a common temporal cue.
The ablation therefore focuses on whether the scheduler observes representation similarity through CKA, robot motion through speed, or both.

\begin{table}[htbp]
  \centering
  \caption{
    \textbf{Effect of scheduler observations on GR00T.}
    Each cell reports success rate (\%, $\uparrow$) / FLOPs speedup ($\uparrow$).
    Progress is included as a shared temporal cue; the comparison isolates CKA, speed, and their combination.
  }
  \label{tab:app_scheduler_observation_ablation}
  \begingroup
  \scriptsize
  \setlength{\tabcolsep}{1pt}
  \renewcommand{\arraystretch}{1.06}
  \begin{tabularx}{\linewidth}{@{}l l l *{6}{>{\centering\arraybackslash}X} >{\centering\arraybackslash}X@{}}
    \toprule
    \multirow{2}{*}{\textbf{SIMPLER}} &
    \multirow{2}{*}{\textbf{Stage}} &
    \multirow{2}{*}{\textbf{Obs.}} &
    \multicolumn{6}{c}{\textbf{Success Rate (\%, $\uparrow$) / FLOPs Speedup ($\uparrow$)}} &
    \multirow{2}{*}{\textbf{AVG}} \\
    \cmidrule(lr){4-9}
    & & & CloseD & MoveNear & OpenD & PickCan & PickObj & PlaceD & \\
    \midrule
    \multirow{6}{*}{\makecell{\textbf{Google}\\\textbf{Robot}}}
    & Baseline & -- & 82.00 / 1.00 & 92.00 / 1.00 & 49.50 / 1.00 & 99.50 / 1.00 & 92.50 / 1.00 & 11.00 / 1.00 & 71.08 / 1.00 \\
    \cmidrule(lr){2-10}
    & Stage1 & CKA+progress & 78.50 / 2.49 & 79.50 / 2.50 & 59.00 / 2.46 & 90.50 / 2.53 & 82.00 / 2.50 & 7.00 / 2.45 & 66.08 / 2.47 \\
    & Stage1 & speed+progress & 79.50 / 2.49 & 78.00 / 2.46 & 67.50 / 2.47 & 90.00 / 2.47 & 84.00 / 2.47 & 8.50 / 2.42 & 67.92 / 2.45 \\
    \cmidrule(lr){2-10}
    & Stage2 & CKA+progress & 80.50 / 2.43 & 79.00 / 2.39 & 60.50 / 2.43 & 91.50 / 2.30 & 81.50 / 2.38 & 7.00 / 2.44 & 66.67 / 2.42 \\
    & Stage2 & speed+progress & 78.50 / \textbf{2.59} & 79.00 / \textbf{2.61} & 63.00 / \textbf{2.58} & 92.00 / \textbf{2.64} & 80.50 / \textbf{2.59} & 7.50 / \textbf{2.55} & 66.75 / \textbf{2.58} \\
    & \cellcolor{ElegantPurple!15}Stage2 & \cellcolor{ElegantPurple!15}\makecell{CKA+speed\\+progress} & \cellcolor{ElegantPurple!15}\textbf{81.00} / 2.39 & \cellcolor{ElegantPurple!15}\textbf{95.50} / 1.90 & \cellcolor{ElegantPurple!15}\textbf{74.50} / 2.31 & \cellcolor{ElegantPurple!15}\textbf{98.00} / 2.35 & \cellcolor{ElegantPurple!15}\textbf{88.50} / 2.59 & \cellcolor{ElegantPurple!15}\textbf{12.50} / 2.39 & \cellcolor{ElegantPurple!15}\textbf{75.00} / 2.35 \\
    \bottomrule
  \end{tabularx}
  \vspace{2pt}
  \begin{tabularx}{\linewidth}{@{}l l l *{7}{>{\centering\arraybackslash}X} >{\centering\arraybackslash}X@{}}
    \toprule
    \multirow{2}{*}{\textbf{SIMPLER}} &
    \multirow{2}{*}{\textbf{Stage}} &
    \multirow{2}{*}{\textbf{Obs.}} &
    \multicolumn{7}{c}{\textbf{Success Rate (\%, $\uparrow$) / FLOPs Speedup ($\uparrow$)}} &
    \multirow{2}{*}{\textbf{AVG}} \\
    \cmidrule(lr){4-10}
    & & & Carrot & CloseD & OpenD & Basket & Sink & Spoon & Stack & \\
    \midrule
    \multirow{6}{*}{\textbf{WidowX}}
    & Baseline & -- & 60.50 / 1.00 & 98.50 / 1.00 & 93.00 / 1.00 & 37.50 / 1.00 & 37.50 / 1.00 & 71.50 / 1.00 & 7.00 / 1.00 & 57.93 / 1.00 \\
    \cmidrule(lr){2-11}
    & Stage1 & CKA+progress & 55.50 / 2.45 & 94.50 / 2.43 & 86.50 / 2.45 & 19.50 / 2.45 & 34.50 / 2.44 & 69.00 / 2.44 & 8.50 / 2.44 & 52.57 / 2.44 \\
    & Stage1 & speed+progress & 53.50 / 2.53 & 94.00 / 2.50 & 79.50 / 2.53 & 18.50 / 2.50 & 29.50 / 2.54 & 62.00 / 2.51 & 8.50 / 2.51 & 49.36 / 2.52 \\
    \cmidrule(lr){2-11}
    & Stage2 & CKA+progress & 53.00 / 2.43 & 93.50 / 2.41 & 83.00 / 2.42 & 19.50 / 2.44 & 33.50 / 2.43 & 58.50 / 2.43 & \textbf{14.00} / 2.43 & 50.71 / 2.43 \\
    & Stage2 & speed+progress & 59.00 / 2.52 & 94.00 / 2.50 & \textbf{87.00} / 2.54 & 19.00 / 2.51 & 37.50 / 2.54 & 68.00 / 2.51 & 8.50 / 2.52 & 53.29 / 2.52 \\
    & \cellcolor{ElegantPurple!15}Stage2 & \cellcolor{ElegantPurple!15}\makecell{CKA+speed\\+progress} & \cellcolor{ElegantPurple!15}\textbf{63.50} / \textbf{2.58} & \cellcolor{ElegantPurple!15}\textbf{99.50} / \textbf{2.54} & \cellcolor{ElegantPurple!15}85.00 / \textbf{2.56} & \cellcolor{ElegantPurple!15}\textbf{31.50} / \textbf{2.52} & \cellcolor{ElegantPurple!15}\textbf{41.00} / \textbf{2.57} & \cellcolor{ElegantPurple!15}\textbf{72.50} / \textbf{2.55} & \cellcolor{ElegantPurple!15}13.50 / \textbf{2.55} & \cellcolor{ElegantPurple!15}\textbf{58.07} / \textbf{2.55} \\
    \bottomrule
  \end{tabularx}
  \endgroup
\end{table}

\paragraph{Analysis.}
CKA+progress gives the scheduler a semantic stability cue, so it can detect when the Vision--LLM representation changes across time.
However, it cannot observe whether the robot is entering contact, moving quickly, or requiring low-level action refinement.
Speed+progress gives the opposite view: it captures motion continuity and often keeps FLOPs speedup high, but it cannot directly detect visual-language changes such as object displacement, drawer-state changes, or instruction-relevant scene transitions.
The full CKA+speed+progress observation combines these complementary cues.
It raises the average success rate to 75.00\% on Google Robot and 58.07\% on WidowX; across all 13 tasks, it improves the average from the reduced-observation range of 57.92--59.50\% to 65.88\% while preserving a 2.47$\times$ FLOPs speedup.
This supports the scheduler design: CKA helps decide when backbone reuse is safe, speed helps decide when action-head reuse is safe, and progress provides a shared temporal prior rather than the main ablated signal.

\newpage
\section{Qualitative Visualizations}
\label{app:qualitative_visualizations}

\subsection{Additional Real-World Rollouts}
\label{app:real_world_rollout_visualizations}

\begin{figure}[H]
    \centering
    \includegraphics[width=\linewidth]{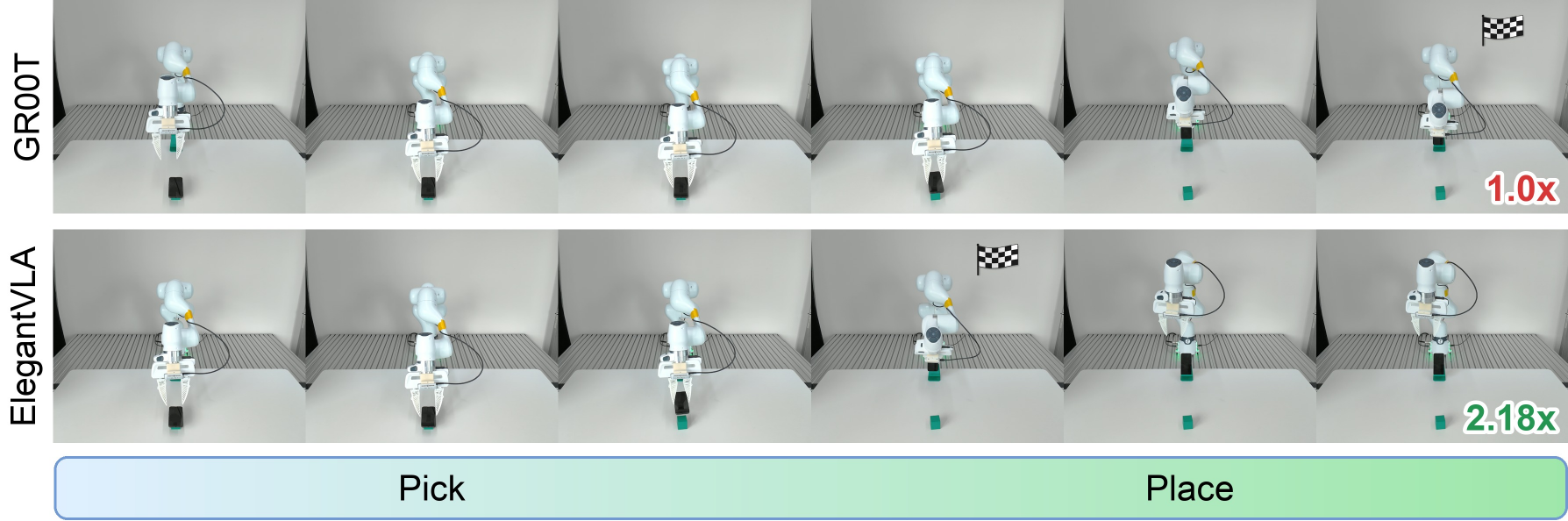}
    \caption{
        \textbf{Real-world rollout on Phone Stand.}
        \sysname preserves precise placement and release on a stationary task, indicating that acceleration does not degrade fine-grained execution.
    }
    \label{fig:app_real_phone_stand_rollout}
\end{figure}

\begin{figure}[H]
    \centering
    \includegraphics[width=\linewidth]{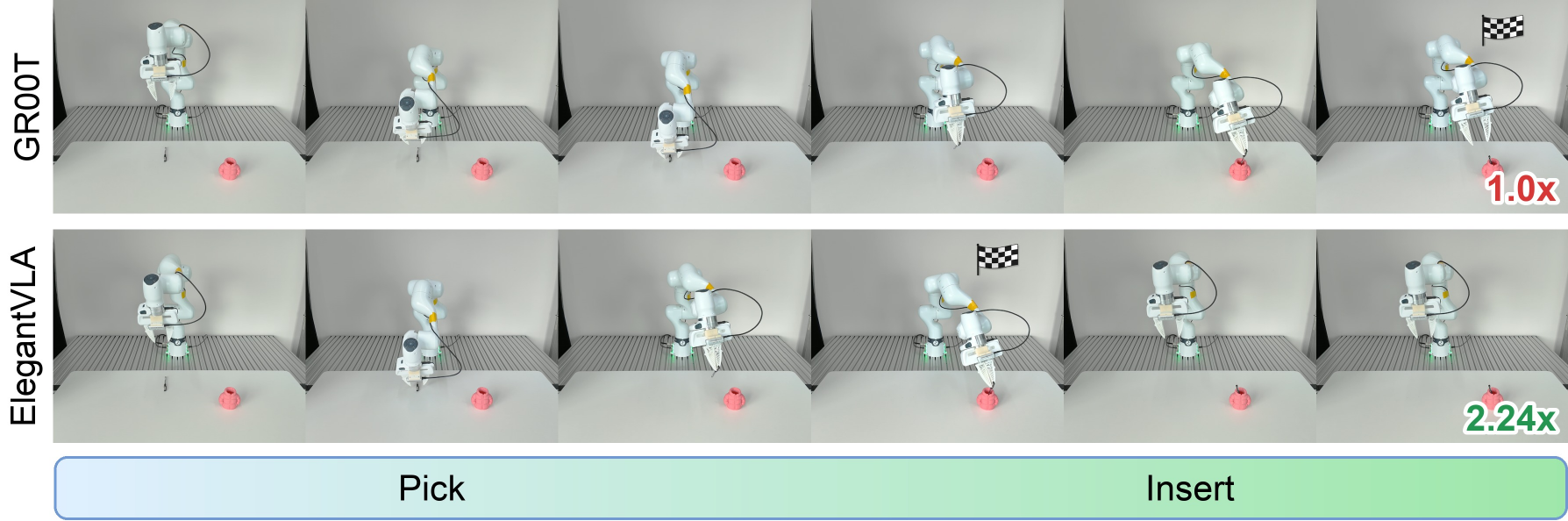}
    \caption{
        \textbf{Real-world rollout on Pen Holder.}
        Adaptive reuse keeps the contact-sensitive insertion phase, providing a qualitative check that temporal reuse does not remove precision-critical behavior.
    }
    \label{fig:app_real_pen_holder_rollout}
\end{figure}

\begin{figure}[H]
    \centering
    \includegraphics[width=\linewidth]{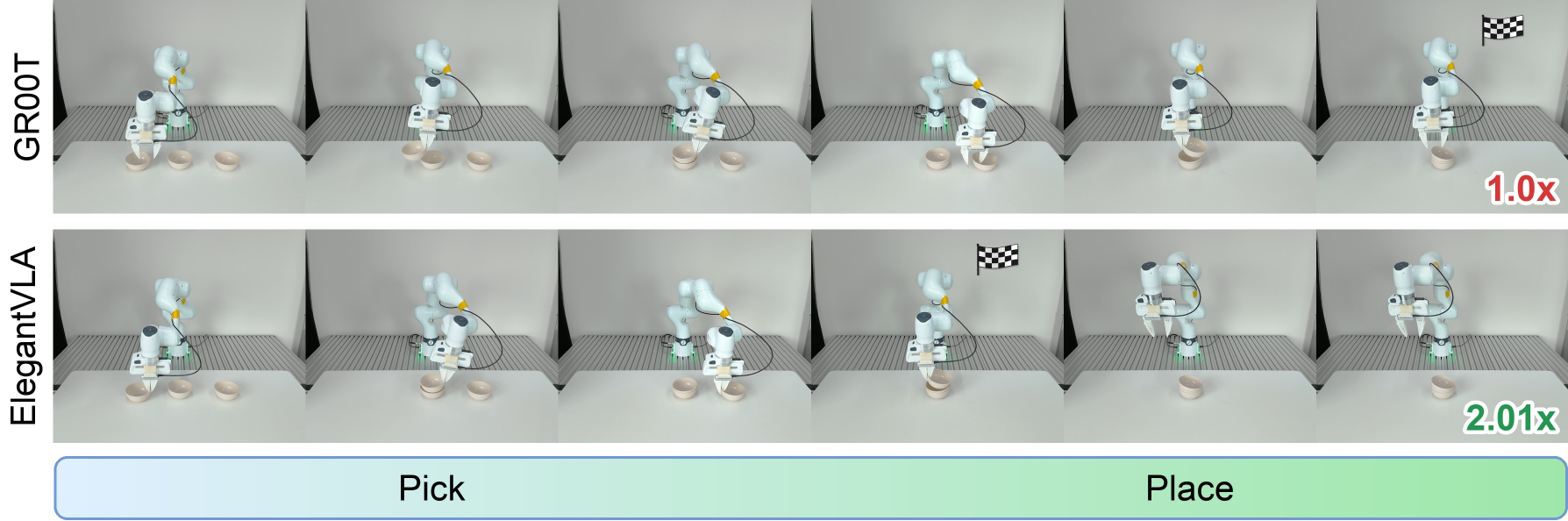}
    \caption{
        \textbf{Real-world rollout on Stack Bowls.}
        The sequence preserves grasping, alignment, and release, showing that acceleration does not collapse the manipulation into a coarse action sequence.
    }
    \label{fig:app_real_stack_bowls_rollout}
\end{figure}

\begin{figure}[H]
    \centering
    \includegraphics[width=\linewidth]{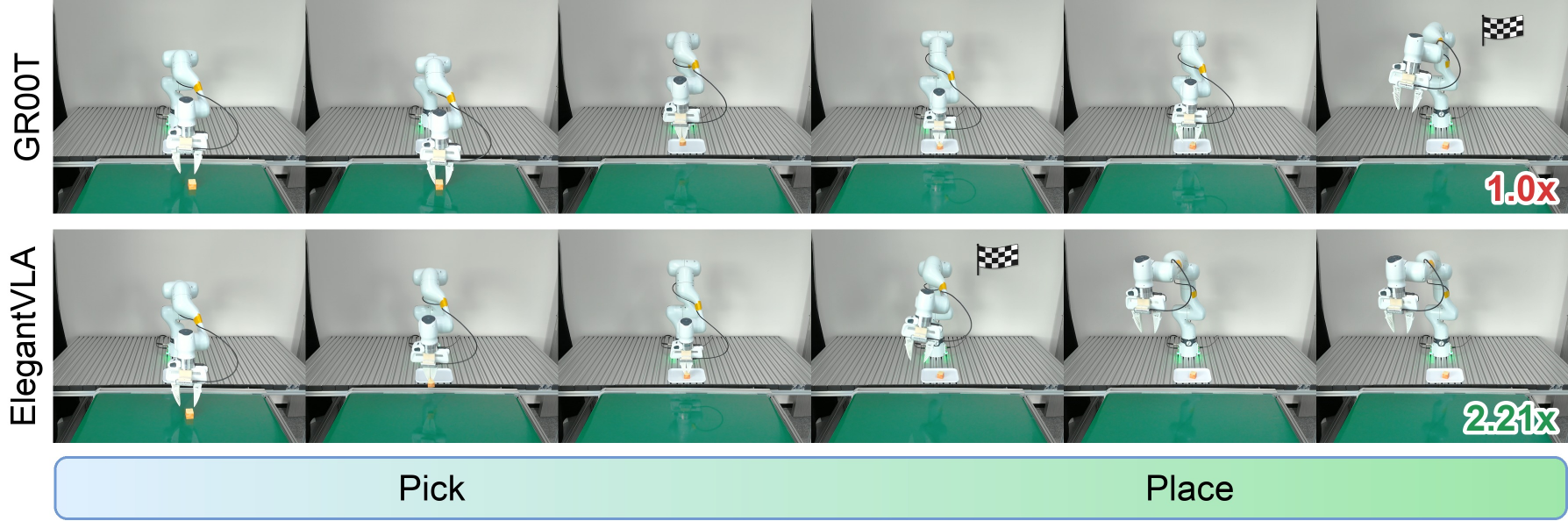}
    \caption{
        \textbf{Real-world rollout on Toast.}
        The conveyor-belt setting tests whether lower latency lets the policy keep correcting the end-effector motion around a moving target.
    }
    \label{fig:app_real_toast_rollout}
\end{figure}

\begin{figure}[H]
    \centering
    \includegraphics[width=\linewidth]{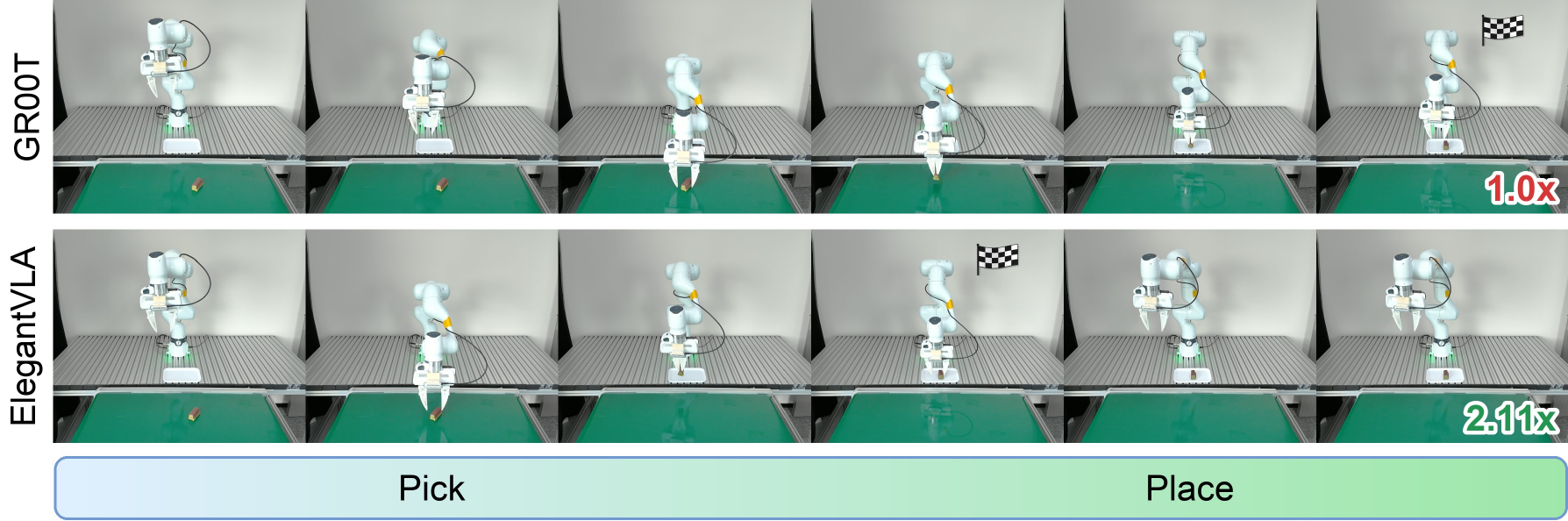}
    \caption{
        \textbf{Real-world rollout on Chocolate Bun.}
        This additional moving-target rollout shows the same responsiveness regime as the main pineapple-bun example, where timely corrections are important.
    }
    \label{fig:app_real_chocolate_bun_rollout}
\end{figure}

To complement the pineapple-bun rollout shown in Fig.~\ref{fig:real_qualitative_execution}, we show the remaining five real-world tasks individually in Figs.~\ref{fig:app_real_phone_stand_rollout}--\ref{fig:app_real_chocolate_bun_rollout}.
These rollouts evaluate two different effects of acceleration in the physical setup.
For stationary manipulation tasks, the goal is to verify that temporal reuse does not sacrifice execution precision.
For conveyor-belt pickup tasks, the goal is to verify that lower inference latency gives the policy more responsive closed-loop behavior while the target is moving.

Phone Stand, Pen Holder, and Stack Bowls are stationary precision tasks.
They require contact-sensitive behaviors such as approach, alignment, insertion or placement, and release.
The qualitative question is therefore not whether the robot simply moves faster, but whether acceleration preserves the fine-grained phases needed for accurate manipulation.
The visual sequences show that \sysname retains these phases rather than collapsing the execution into a coarse approach-and-release pattern.
This supports the trial-level results in Appendix~\ref{app:real_world}, where \sysname preserves success on stationary real-world tasks while reducing redundant computation.

Toast and Chocolate Bun, together with the pineapple-bun example in the main paper, stress a different regime.
Because the object pose changes continuously on the conveyor belt, delayed actions make the end-effector chase a stale target state.
Here, acceleration is expected to improve responsiveness: the policy can observe the moving object and issue corrective actions with less observation-action delay.
The rollouts show more target-centered corrections during approach and grasping, indicating that the compute savings translate into a more sensitive closed-loop response rather than only a lower FLOPs count.
Across both task families, the real-world visualizations support the same conclusion as the quantitative breakdown: \sysname preserves precision when the scene is stable and improves reaction to target motion when the scene changes during execution.

\clearpage

\subsection{Simulation Rollouts and Scheduler Behavior}
\label{app:simulation_rollout_visualizations}

The simulation rollouts visualize the task-complexity adaptivity of \sysname.
Rather than applying a fixed lightweight policy throughout an episode, the scheduler changes the amount and type of computation as the task state evolves.
During simple and stable phases, such as smooth transport or visually unchanged approach, the policy can use more aggressive reuse.
When the task becomes more complex, such as object contact, drawer-state change, grasp adjustment, or final placement, the scheduler reduces reuse and allocates more fresh computation.

This adaptation happens at two semantic levels.
For the Vision--LLM backbone, the scheduler mainly uses CKA to estimate high-level semantic stability.
A high CKA value indicates that the current visual-language representation remains close to the cached representation, so backbone reuse is likely safe.
A low CKA value suggests that the task semantics may have changed, for example when the object pose, drawer state, or spatial relation changes, and the scheduler therefore refreshes the high-level representation.
For the action head, the scheduler mainly uses robot speed to estimate low-level control complexity.
Smooth and continuous motion usually corresponds to a stable control phase where action-head reuse is sufficient.
In contrast, rapid speed changes often occur near contact, correction, or release, where fine-grained control matters and aggressive reuse can be risky.

\begin{figure}[H]
    \centering
    \includegraphics[width=\linewidth]{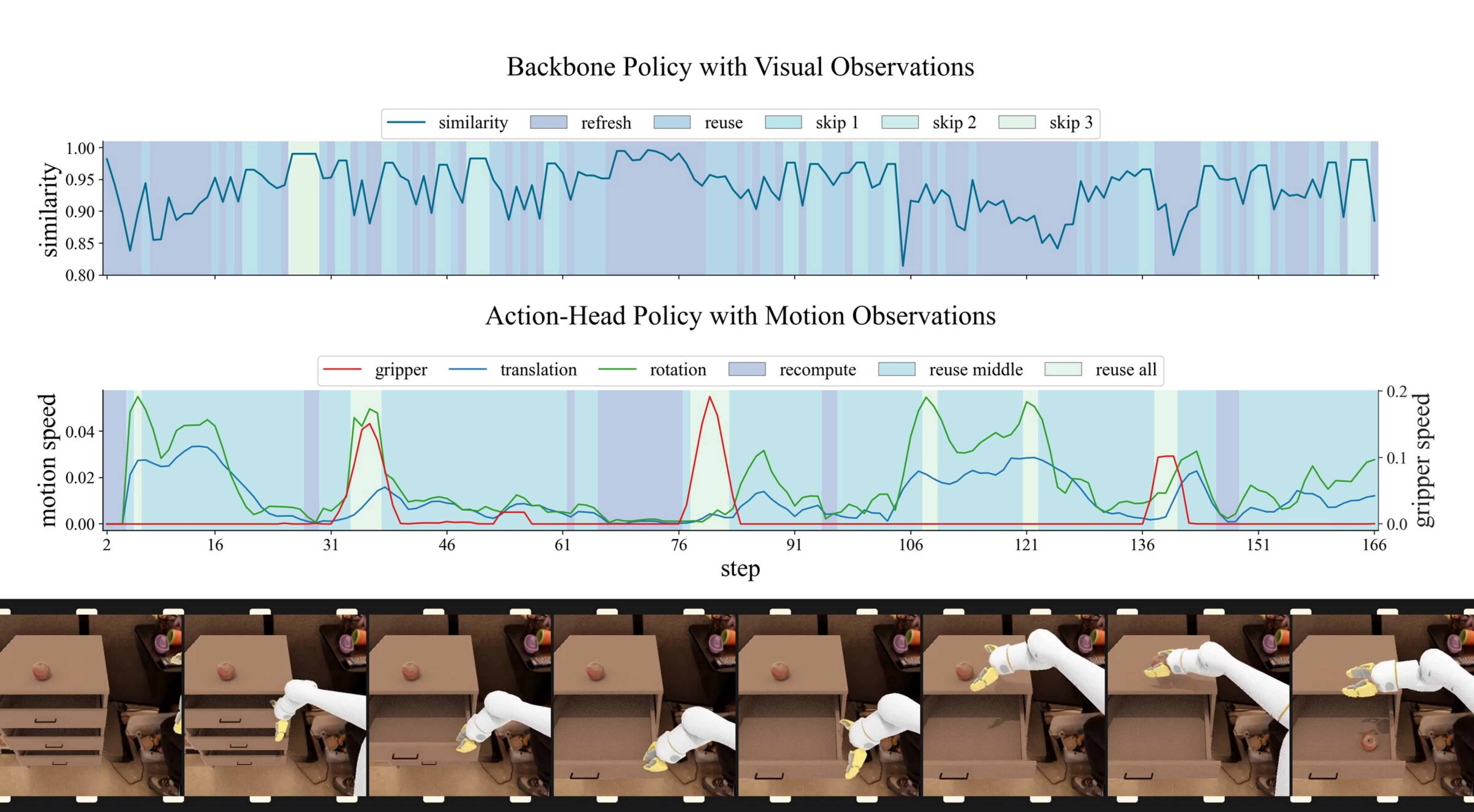}
    \caption{
        \textbf{Representative rollout-level scheduling diagnostic.}
        The background colors show selected execution modes for the Vision--LLM backbone and the action head, while the curves show CKA-based representation similarity and robot motion speeds along the same trajectory.
        The backbone follows high-level semantic stability, whereas the action head follows low-level control stability.
    }
    \label{fig:app_sim_scheduler_diagnostic}
\end{figure}

\begin{figure}[H]
    \centering
    \includegraphics[width=\linewidth]{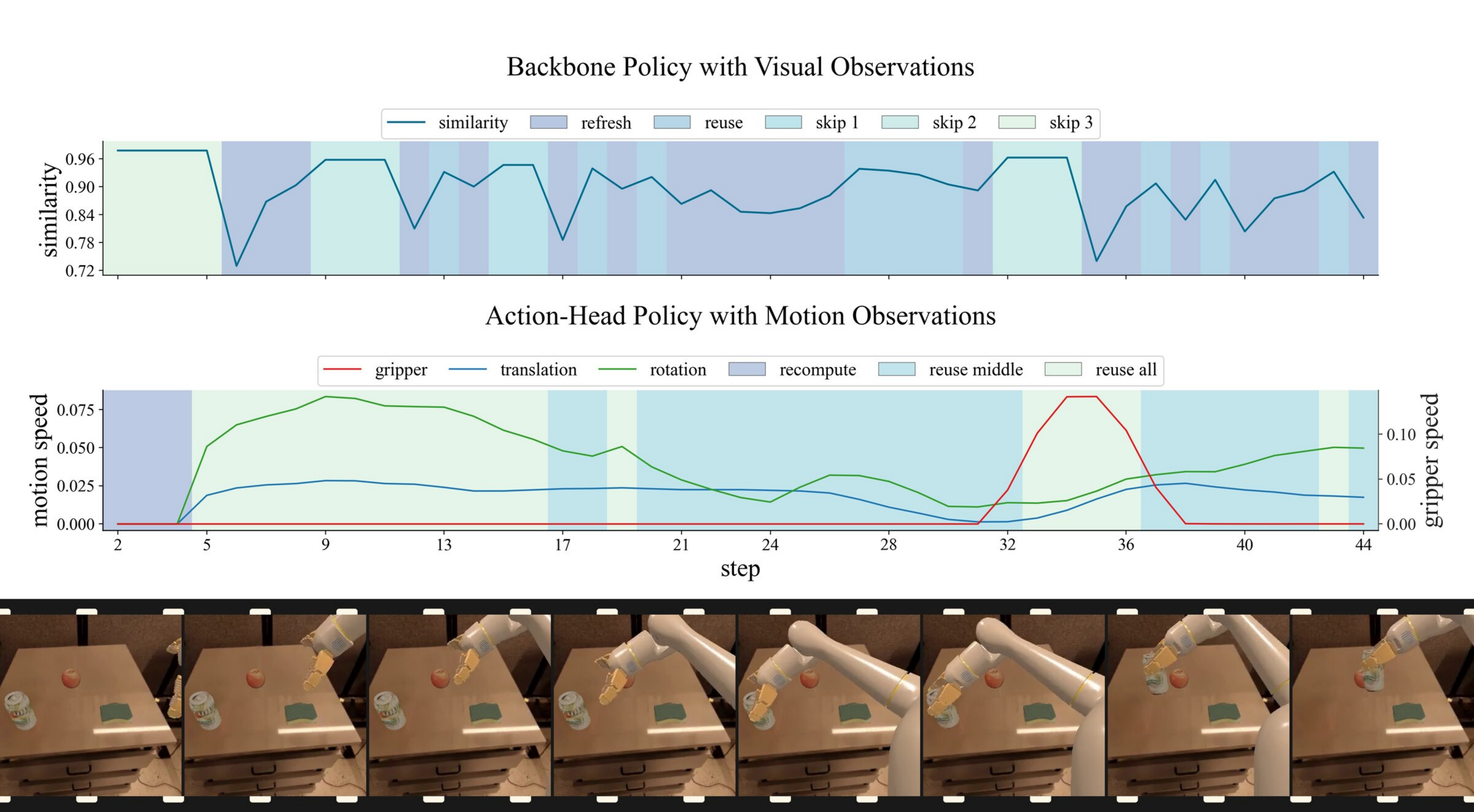}
    \caption{
        \textbf{Google Robot simulation rollout on Move Near.}
        The spatial transport task shows how the scheduler changes lightweight modes as high-level task complexity changes.
    }
    \label{fig:app_google_move_near_rollout}
\end{figure}

\begin{figure}[H]
    \centering
    \includegraphics[width=\linewidth]{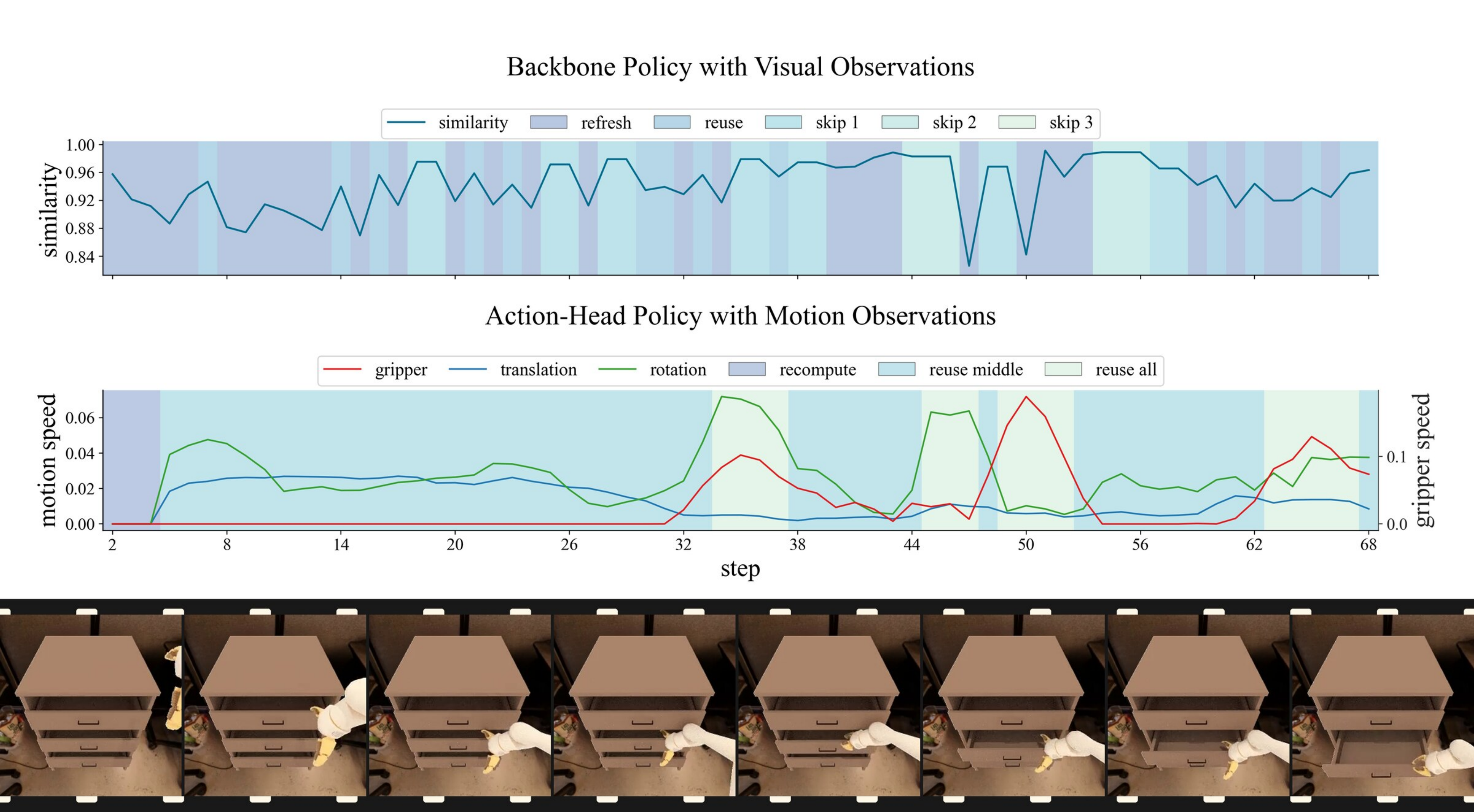}
    \caption{
        \textbf{Google Robot simulation rollout on Open Drawer.}
        The drawer manipulation task illustrates that \sysname adapts computation to semantic state changes rather than following one fixed reuse pattern.
    }
    \label{fig:app_google_open_drawer_rollout}
\end{figure}

\begin{figure}[H]
    \centering
    \includegraphics[width=\linewidth]{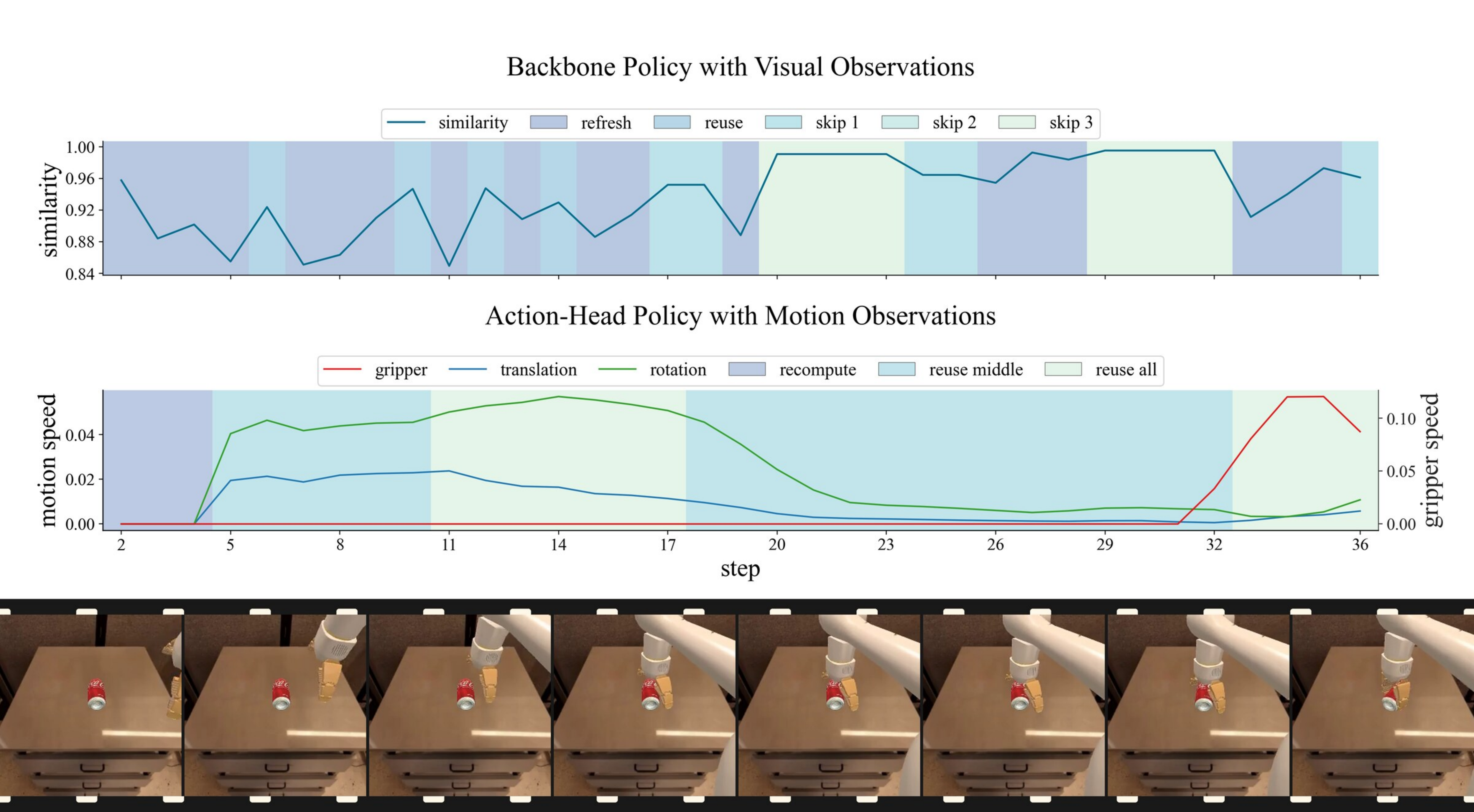}
    \caption{
        \textbf{Google Robot simulation rollout on Pick Coke Can.}
        The object pickup task shows how high-level object state and low-level grasp control jointly affect the scheduler's lightweight choices.
    }
    \label{fig:app_google_pick_can_rollout}
\end{figure}

\begin{figure}[H]
    \centering
    \includegraphics[width=\linewidth]{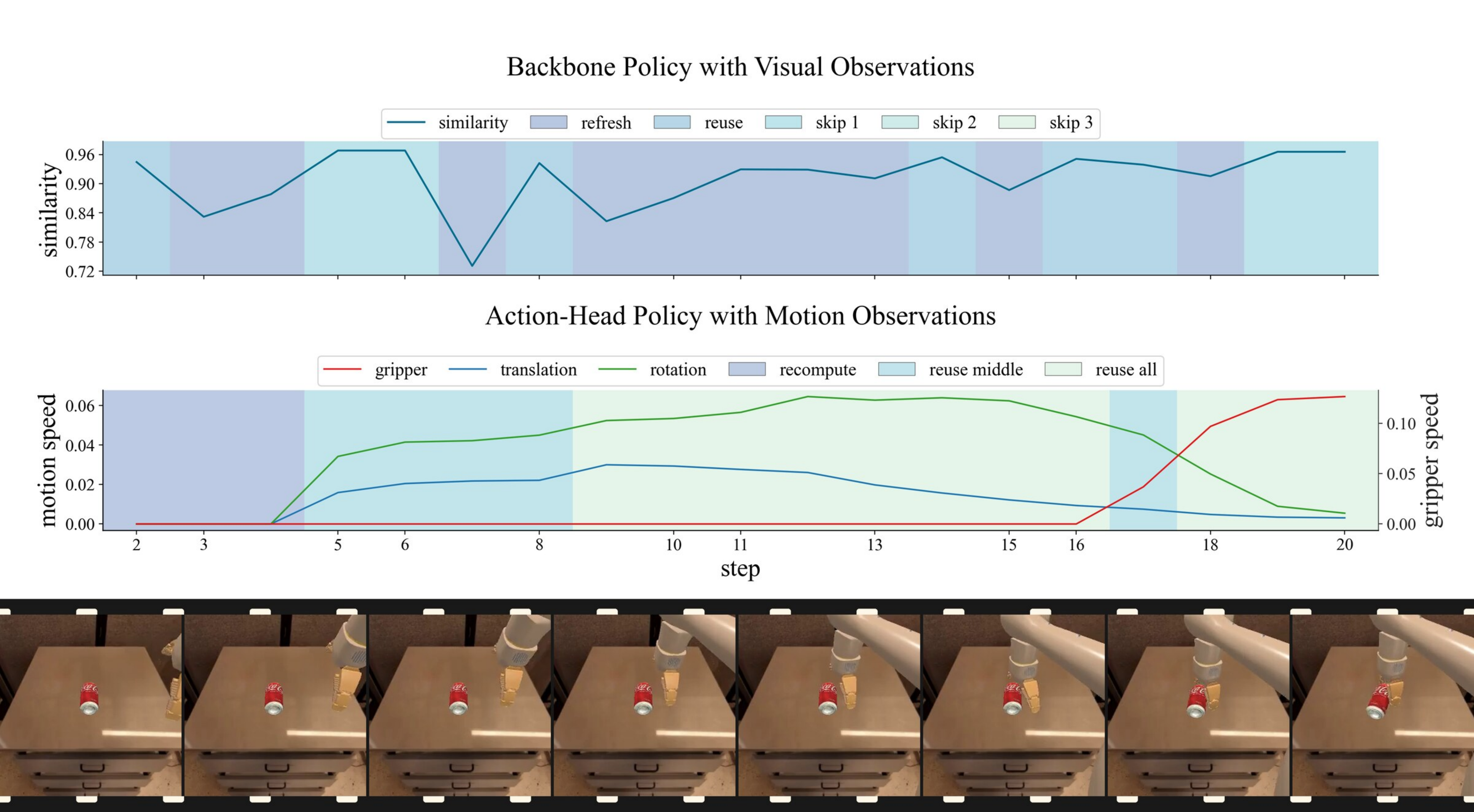}
    \caption{
        \textbf{Google Robot simulation rollout on Pick Object.}
        The generic pickup task provides another example where \sysname adjusts computation as semantic and control complexity vary over time.
    }
    \label{fig:app_google_pick_object_rollout}
\end{figure}

\begin{figure}[H]
    \centering
    \includegraphics[width=\linewidth]{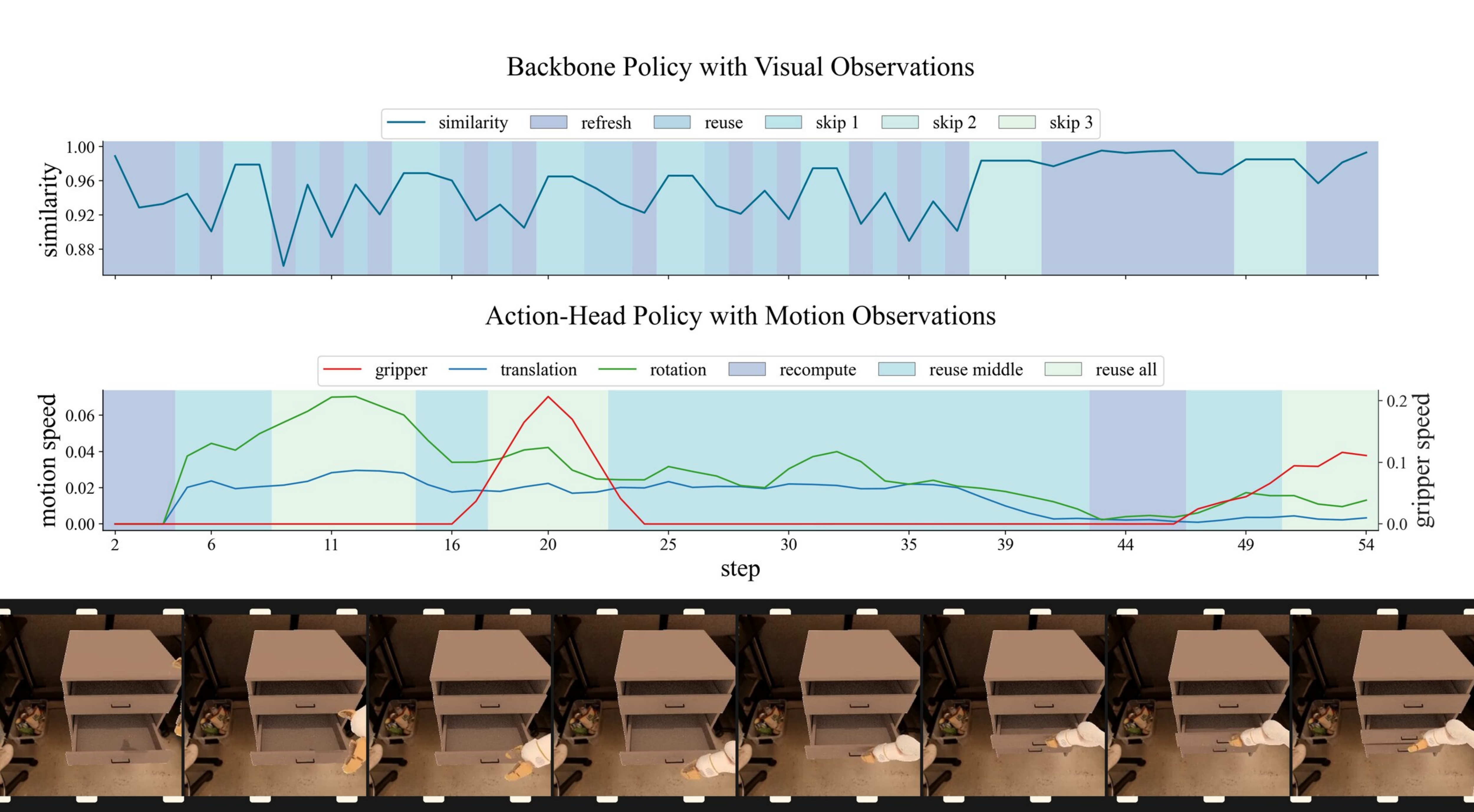}
    \caption{
        \textbf{Google Robot simulation rollout on Close Drawer.}
        Closing the drawer requires the scheduler to preserve computation around drawer-state changes while allowing reuse during stable motion.
    }
    \label{fig:app_google_close_drawer_rollout}
\end{figure}

\begin{figure}[H]
    \centering
    \includegraphics[width=\linewidth]{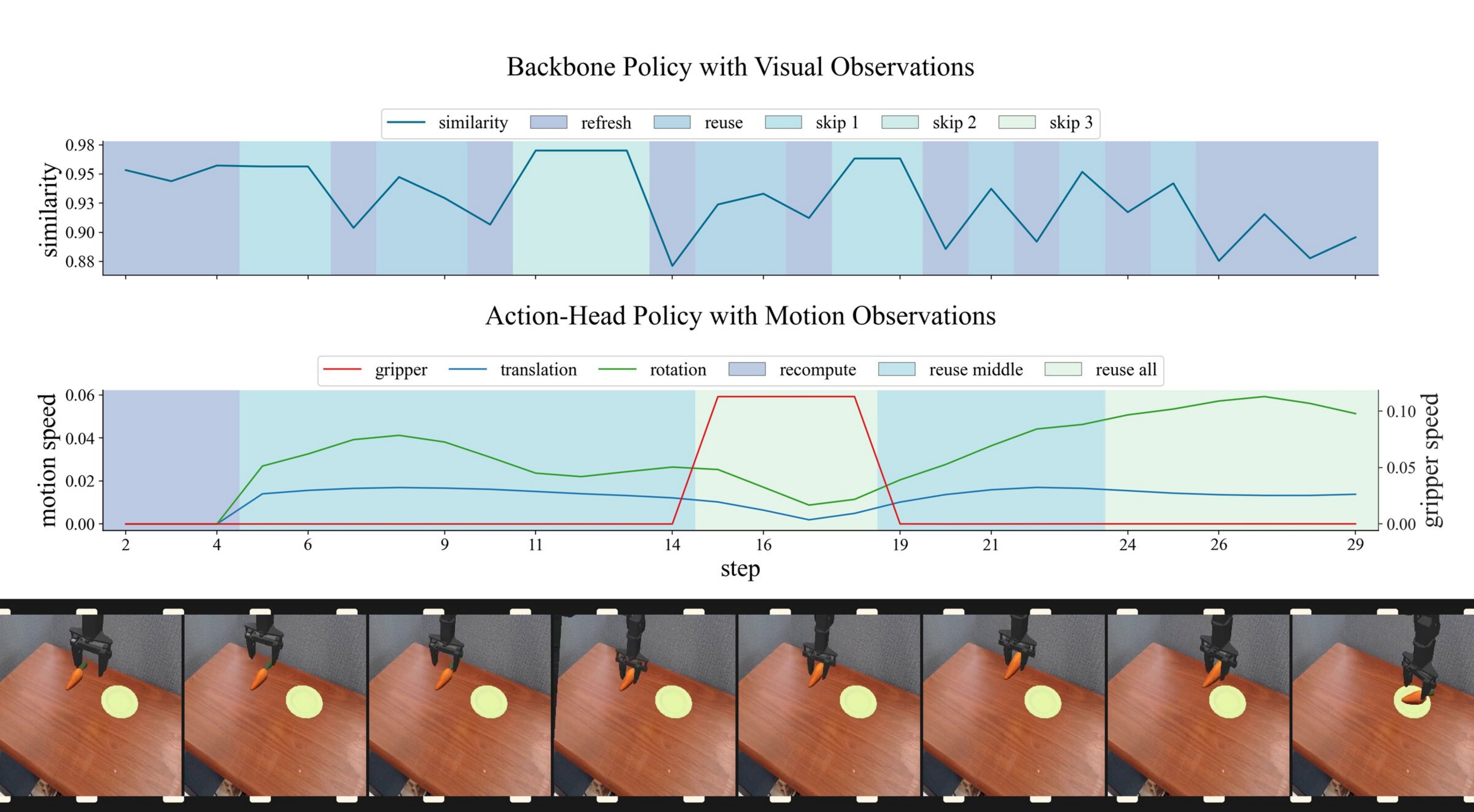}
    \caption{
        \textbf{WidowX Robot simulation rollout on Carrot on Plate.}
        The placement task tests whether the two-level scheduler remains adaptive under the WidowX visual domain.
    }
    \label{fig:app_widowx_carrot_plate_rollout}
\end{figure}

\begin{figure}[H]
    \centering
    \includegraphics[width=\linewidth]{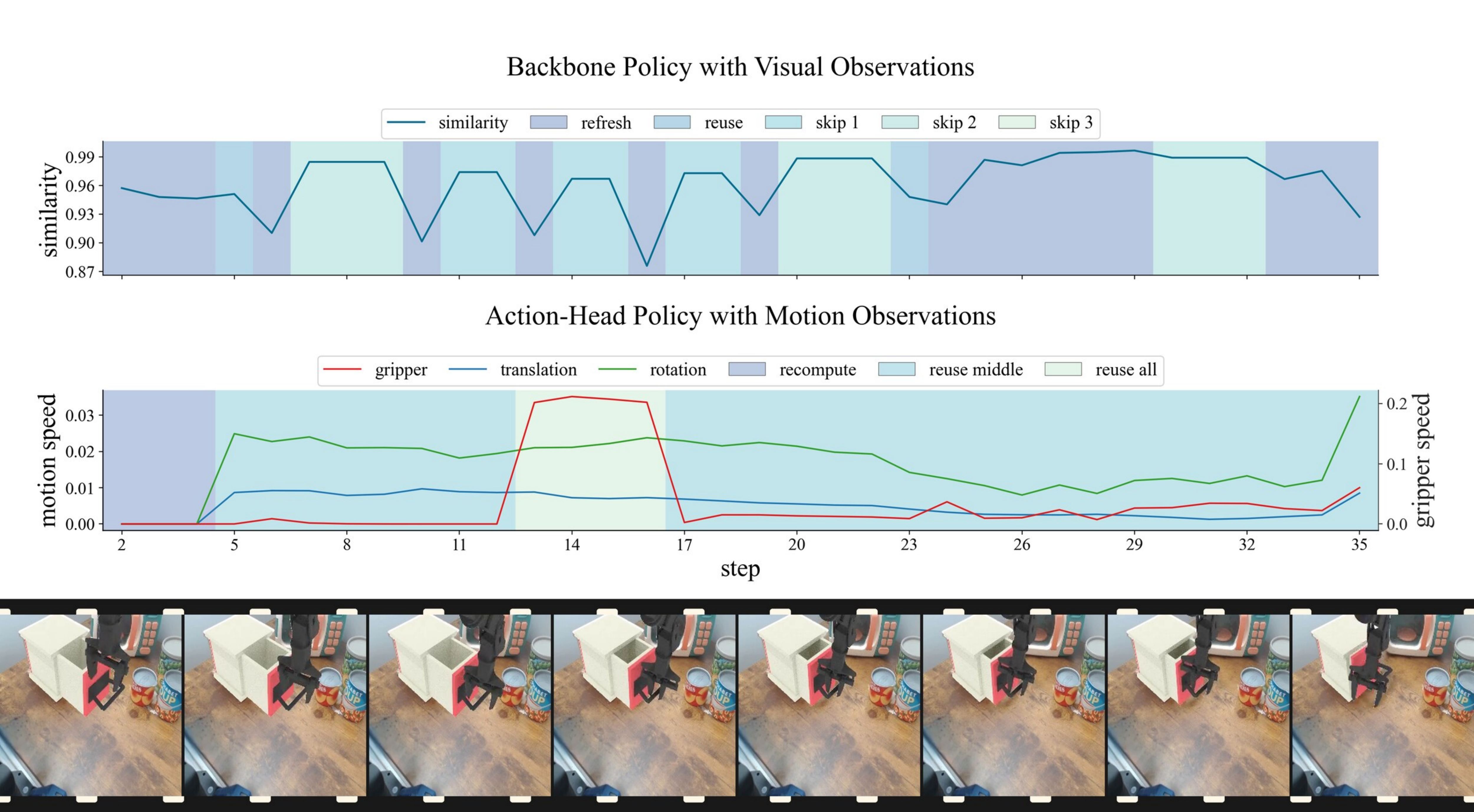}
    \caption{
        \textbf{WidowX Robot simulation rollout on Close Drawer.}
        The contact-rich drawer interaction shows how low-level control complexity affects action-head reuse.
    }
    \label{fig:app_widowx_close_drawer_rollout}
\end{figure}

\begin{figure}[H]
    \centering
    \includegraphics[width=\linewidth]{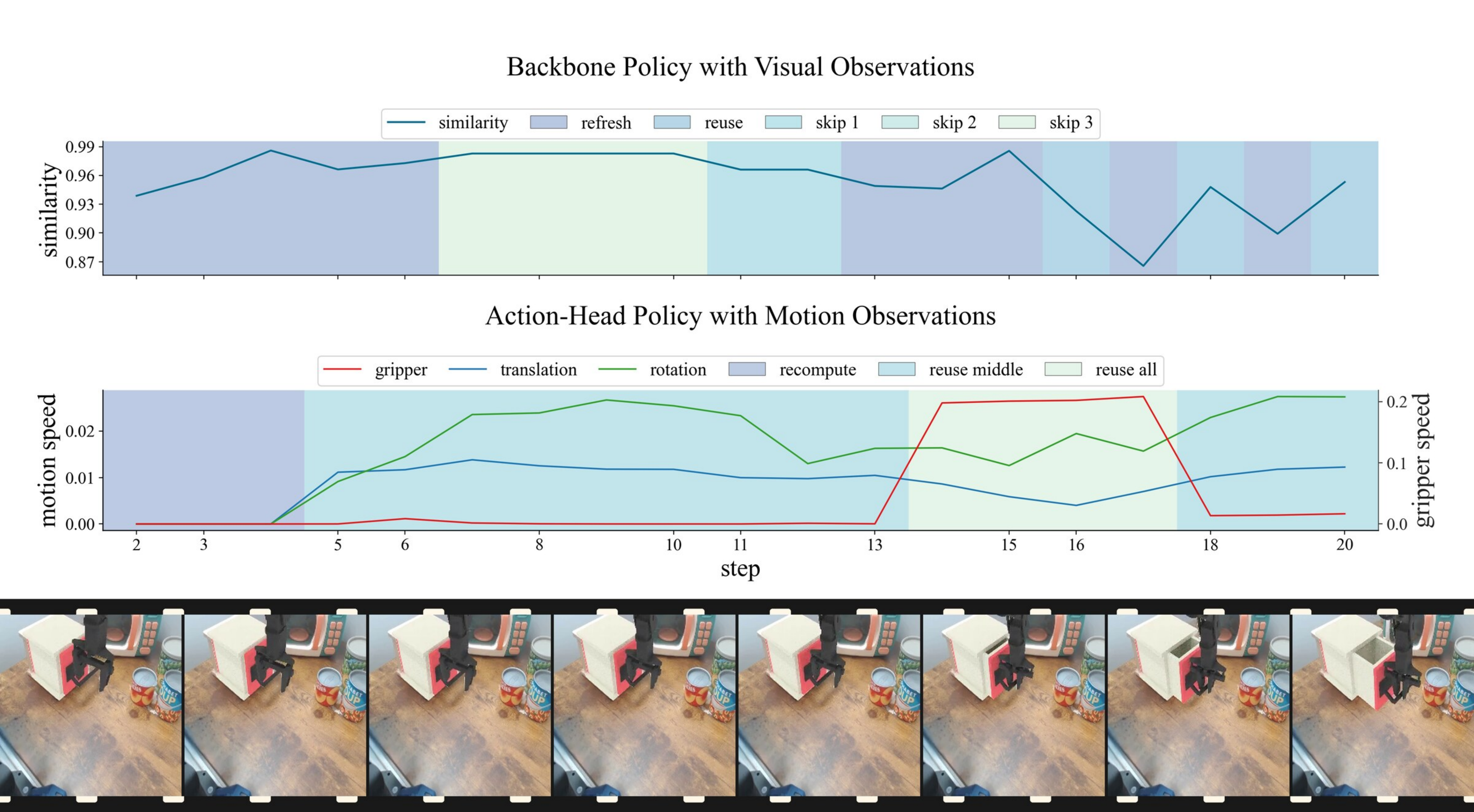}
    \caption{
        \textbf{WidowX Robot simulation rollout on Open Drawer.}
        Opening the drawer provides another contact-sensitive example where the scheduler should reduce aggressive reuse around physical interaction.
    }
    \label{fig:app_widowx_open_drawer_rollout}
\end{figure}

\begin{figure}[H]
    \centering
    \includegraphics[width=\linewidth]{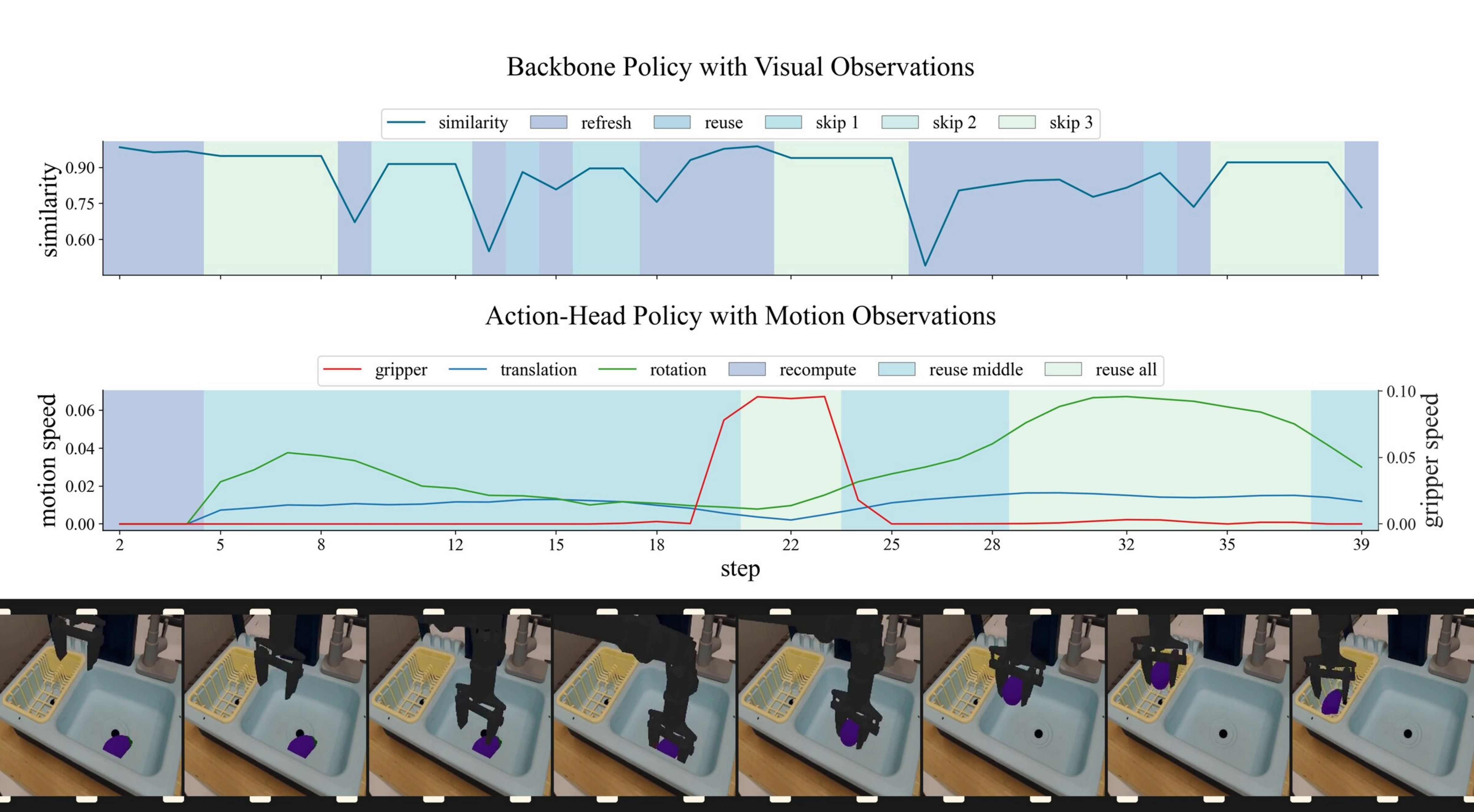}
    \caption{
        \textbf{WidowX Robot simulation rollout on Put Eggplant in Basket.}
        The final placement changes the required object configuration, so the scheduler preserves computation during the more complex control phase.
    }
    \label{fig:app_widowx_eggplant_basket_rollout}
\end{figure}

\begin{figure}[H]
    \centering
    \includegraphics[width=\linewidth]{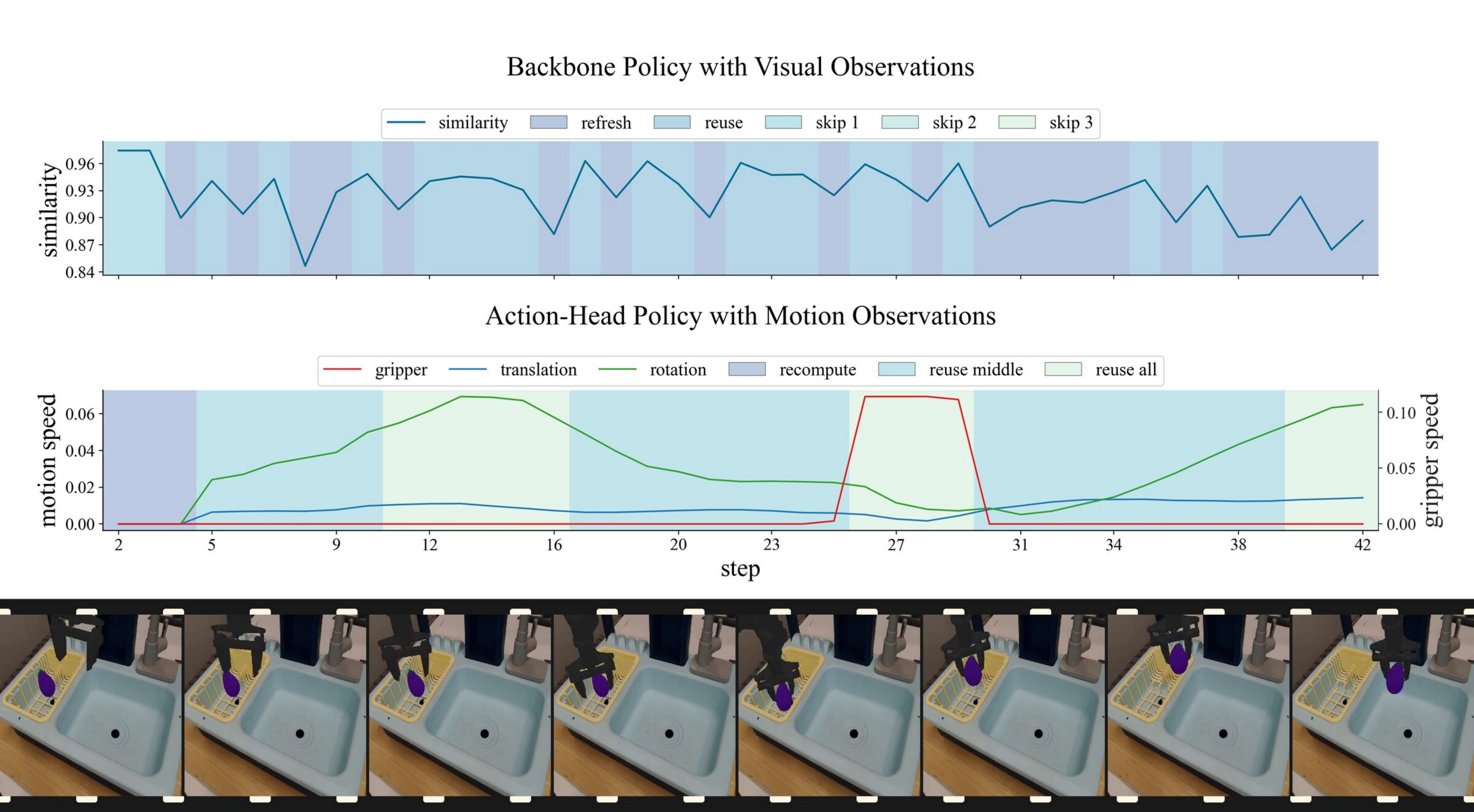}
    \caption{
        \textbf{WidowX Robot simulation rollout on Put Eggplant in Sink.}
        This placement task further shows task-complexity-adaptive computation under changing final object configurations.
    }
    \label{fig:app_widowx_eggplant_sink_rollout}
\end{figure}

\begin{figure}[H]
    \centering
    \includegraphics[width=\linewidth]{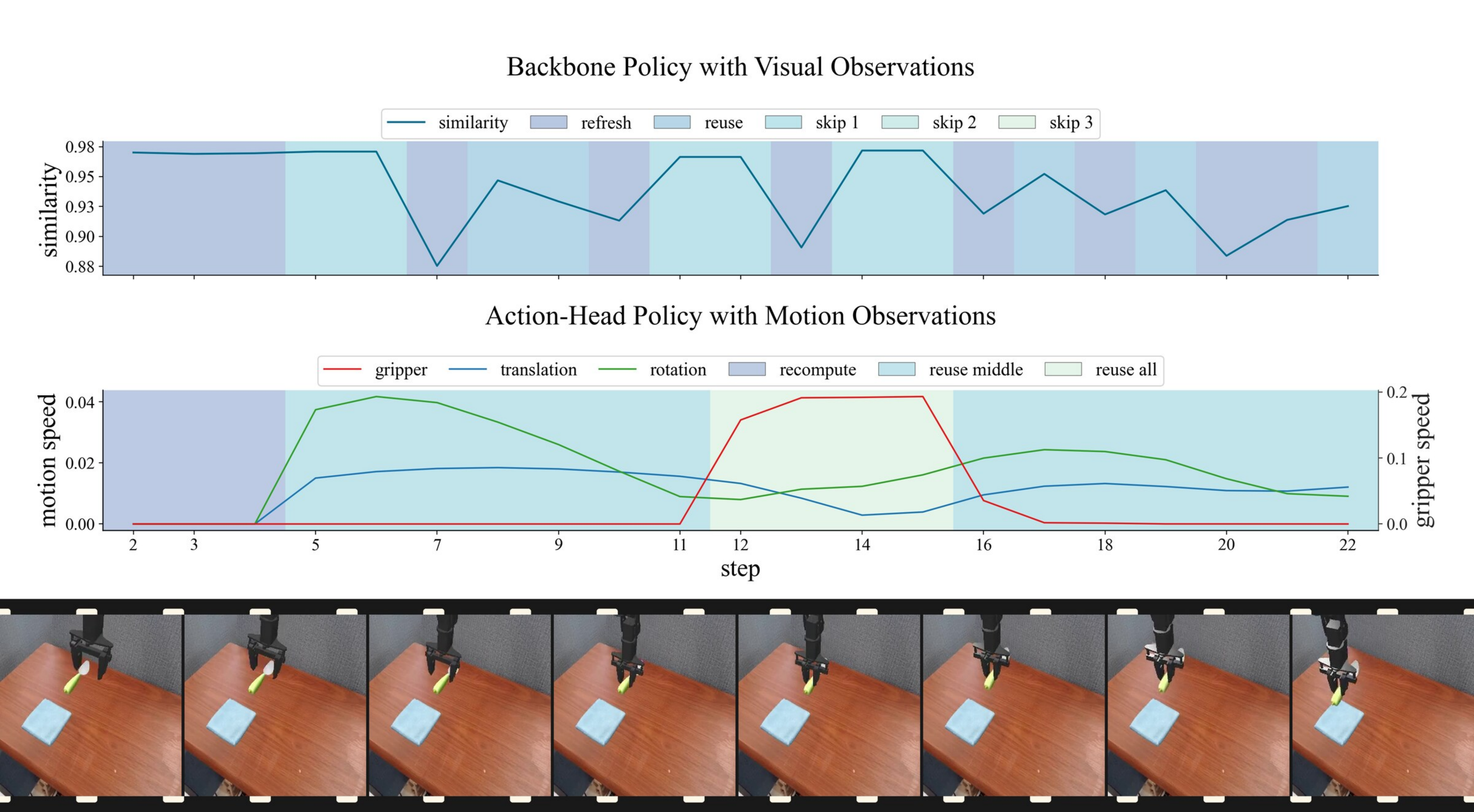}
    \caption{
        \textbf{WidowX Robot simulation rollout on Spoon on Towel.}
        The scheduler keeps lightweight inference for stable motion while preserving computation near placement and release.
    }
    \label{fig:app_widowx_spoon_towel_rollout}
\end{figure}

\begin{figure}[H]
    \centering
    \includegraphics[width=\linewidth]{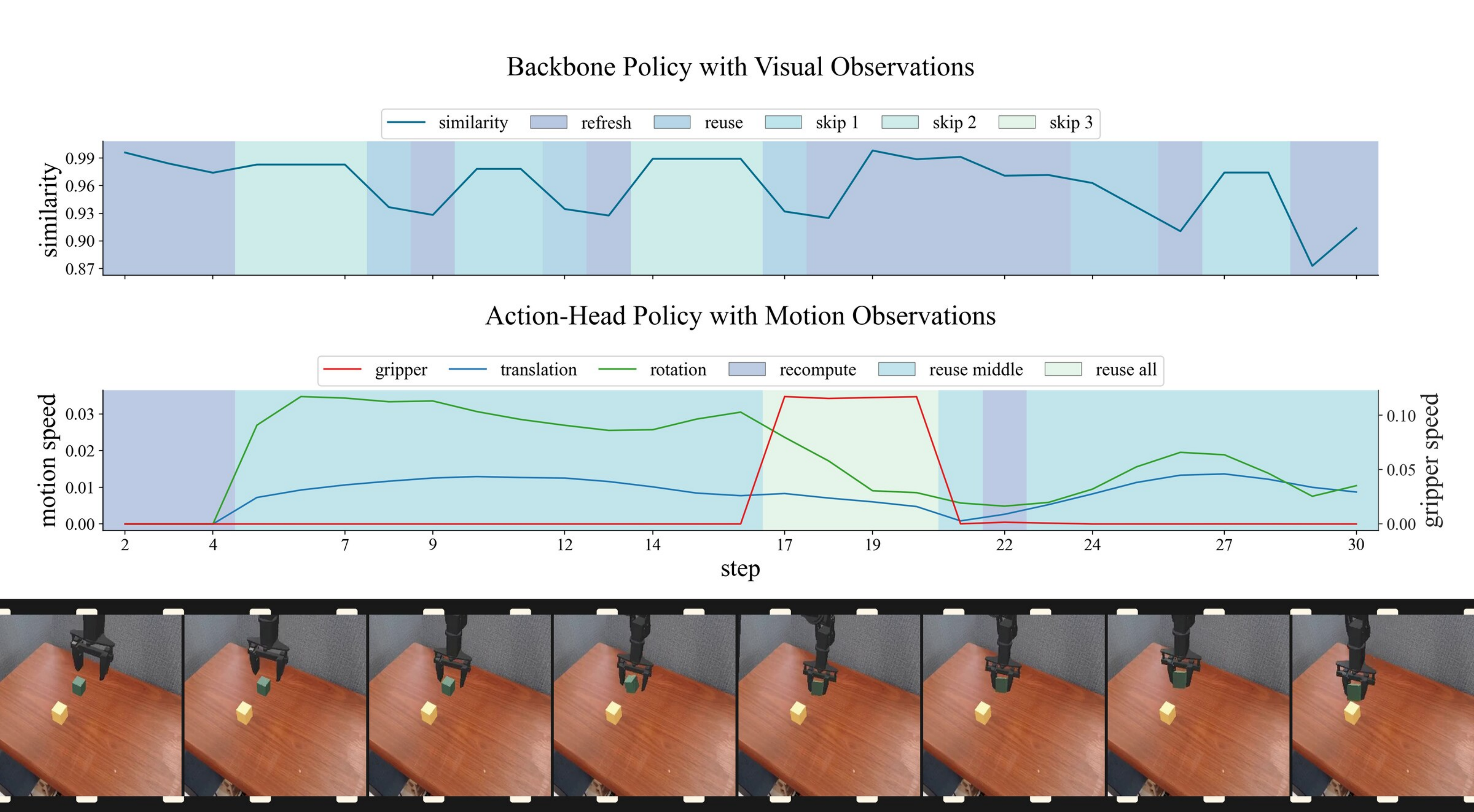}
    \caption{
        \textbf{WidowX Robot simulation rollout on Stack Cube.}
        The stacking task provides an additional contact-sensitive case where low-level alignment and release require less aggressive action-head reuse.
    }
    \label{fig:app_widowx_stack_cube_rollout}
\end{figure}

Fig.~\ref{fig:app_sim_scheduler_diagnostic} illustrates this two-level adaptation on a representative rollout.
The remaining simulation rollouts in Figs.~\ref{fig:app_google_move_near_rollout}--\ref{fig:app_widowx_stack_cube_rollout} show the same principle across Google Robot and WidowX Robot tasks.
Across different task demands, \sysname uses lightweight inference during easy phases while preserving computation for semantically or physically complex phases.
This behavior is the key qualitative evidence that our acceleration is task-complexity adaptive rather than a fixed reduction in model computation.

\clearpage

\end{document}